%% file: main.tex
\documentclass{article}

\usepackage[preprint]{neurips_2025}

\usepackage[pagebackref,colorlinks,allcolors=nipsblue]{hyperref}
\usepackage[table]{xcolor}
\usepackage{multirow}
\usepackage{array}
\usepackage{graphicx}
\usepackage{makecell}
\usepackage[none]{hyphenat}
\usepackage[utf8]{inputenc} 
\usepackage[T1]{fontenc}    
\usepackage{url}            
\usepackage{booktabs}       
\usepackage{amsfonts}       
\usepackage{nicefrac}       
\usepackage{microtype}      
\usepackage{natbib}
\usepackage{tcolorbox}
\usepackage{enumitem}

\definecolor{aliceblue}{rgb}{0.54, 0.77, 1.0}
\definecolor{nipsblue}{rgb}{0.35,0.49,0.74}

\usepackage{graphicx}
\usepackage{wrapfig}
\usepackage{amsmath}
\usepackage{subfig}

\usepackage{pifont}     
\usepackage{xcolor}     
\usepackage{hyperref}
\newcommand{\cmark}{\textcolor{green}{\ding{51}}}  
\newcommand{\xmark}{\textcolor{red}{\ding{55}}}    

\title{Towards Robust Evaluation of STEM Education:\\ Leveraging MLLMs in Project-Based Learning}

\author{Xinyi Wu\textsuperscript{1},
        Yanhao Jia\textsuperscript{2},
        Qinglin Zhang\textsuperscript{1},
       Yiran Qin\textsuperscript{3},
       Luwei Xiao\textsuperscript{2},
       Shuai Zhao\textsuperscript{2}\thanks{\quad Corresponding author; shuai.zhao@ntu.edu.sg}\\
{ 
\textsuperscript{1} Shanghai Jiao Tong University, Shanghai, China.
}\vspace{-0.1mm} \\
{ 
\textsuperscript{2} Nanyang Technological University, Singapore.
}\vspace{-0.1mm}\\
{ 
\textsuperscript{3} Shanghai AI Laboratory, Shanghai, China.
}\\
}

\begin{document}

\maketitle

\input{sec/0_abstract}
\input{sec/1_intro}

\input{sec/2_method}

\input{sec/3_experiment}
\input{sec/4_related}
\input{sec/5_conclusion}

\clearpage
{
    \small
    \bibliographystyle{plain}
    \bibliography{reference}
}
\newpage
\input{sec/6.5_appendix}



\end{document}

%% file: sec/0_abstract.tex
\begin{abstract}

Project-Based Learning (PBL) involves a variety of highly correlated multimodal data, making it a vital educational approach within STEM disciplines.
With the rapid development of multimodal large language models (MLLMs), researchers have begun exploring their potential to enhance tasks such as information retrieval, knowledge comprehension, and data generation in educational settings.
However, existing benchmarks fall short in providing both a free-form output structure and a rigorous human expert validation process, limiting their effectiveness in evaluating real-world educational tasks.
Additionally, few methods have developed automated pipelines to assist with the complex responsibilities of teachers leveraging MLLMs, largely due to model hallucination and instability, which lead to unreliable implementation.
To address this gap, we introduce \textbf{PBLBench}, a novel benchmark designed to evaluate complex reasoning grounded in domain-specific knowledge and long-context understanding, thereby challenging models with tasks that closely resemble those handled by human experts.
We also build a new dataset, \textbf{PBL-STEM}, for this complex scenario, which contains over 500 projects with different modalities and multi-disciplinary contexts.
To establish reliable ground truth, we adopt the Analytic Hierarchy Process (AHP), utilizing expert-driven pairwise comparisons to derive structured and weighted evaluation criteria.
We assess the performance of 15 leading MLLMs/LLMs using PBLBench and demonstrate that even the most advanced models achieve only 59\% rank accuracy, underscoring the significant challenges presented by this benchmark.
We believe PBLBench will serve as a catalyst for the development of more capable AI agents, ultimately aiming to alleviate teacher workload and enhance educational productivity.

\end{abstract}

%% file: sec/1_intro.tex
\section{Introduction}
In recent years, the integration of Artificial Intelligence (AI) in education has opened new avenues to enhance instructional methods and streamline assessment practices, particularly within the STEM (Science, Technology, Engineering, and Mathematics) disciplines~\cite{hwang2020vision,li2024alleviating,jia2025uni}.
STEM education, with its emphasis on inquiry, problem-solving, and real-world applications, has increasingly adopted Project-Based Learning (PBL) as a pedagogical approach to cultivate critical thinking and innovation among students~\cite{firdausih2024literature}.
However, the evaluation of PBL projects—characterized by diverse outputs such as research reports, design schematics, code, experimental data, and demo videos significant challenges in maintaining consistency, efficiency, and objectivity in grading.

\begin{figure}[!t]
    \centering
    \includegraphics[width=\linewidth]{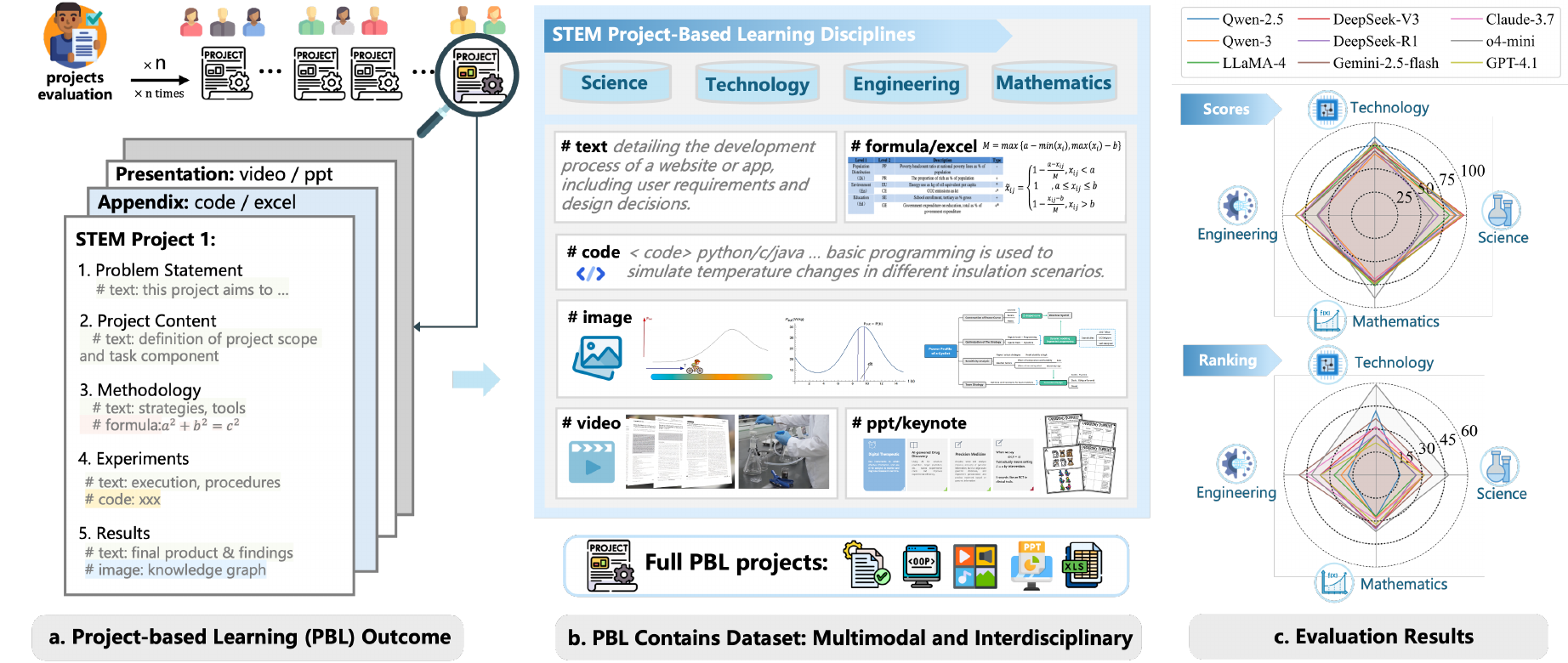}
    \vspace{-5mm}
    \caption{Schematic illustration of the PBL reviewing challenge with different representations.}
    \vspace{-8mm}
    \label{fig:challenge}
\end{figure}

Multimodal large language models (MLLMs)~\cite{nguyen2024kdmcse,nguyen2025enhancing}, which bridge the gap between natural language and other modalities, achieve state-of-the-art performance on several multimodal tasks.
Their potential for cross-modal analysis, reasoning, and evaluation of lengthy and complex narratives makes them promising candidates for supporting teacher assessments in the PBL context, as shown in Figure \ref{fig:challenge}.
Nevertheless, current research on MLLMs has primarily focused on conventional multimodal tasks and general educational assessments, falling short in providing a rigorous human expert validation process.
In addition, due to hallucinations and instability in MLLMs, there are limitations in developing automated pipelines to assist with the complex responsibilities of teachers.
This creates a gap in their application for the multifaceted evaluation of STEM projects, which lacks a unified evaluation framework that combines multiple modalities, such as textual and visual.
Especially, there are hardly any studies exploring whether MLLMs are capable of handling the task related to PBL, which may involve cross-disciplinary knowledge and long-context understanding.

To bridge this gap and comprehensively assess the capabilities of MLLMs for PBL tasks, we propose constructing a multimodal, long-context dataset named \textbf{PBL-STEM}, specifically tailored to STEM-based PBL outcomes.
This dataset aims to capture the inherent modal complexity and richness of STEM projects by integrating a variety of PBL outcomes, including extensive textual descriptions, diagrams, experimental results, code, and other visual content.
Such comprehensive multi-modal data is essential for training and benchmarking AI systems to perform nuanced evaluations.
To ensure alignment with the intricate evaluation criteria typically employed by human experts in STEM education, we utilize the Analytic Hierarchy Process (AHP).
This method involves human experts performing pairwise comparisons of evaluation indicators, thereby constructing the judgment matrix and calculating the relative weights of each indicator.
Following a consistency check, the formulation of evaluation criteria suitable for PBL is finalized.

Our study is positioned at the confluence of AI for education, advanced multimodal modeling, and STEM-oriented PBL.
By developing a specialized benchmark, named \textbf{PBLBench}, we seek to evaluate the performance of state-of-the-art MLLMs in assessing multi-modal STEM-based PBL outcomes.
This benchmark will measure not only the MLLMs’ ability to process long-context inputs but also their proficiency in handling cross-modal information critical for a complete understanding of STEM projects.
Furthermore, the benchmark is designed to reflect real-world scoring and ranking scenarios, where AI-assisted assessment could significantly reduce teachers' workload and provide prompt, constructive feedback to students.

We conduct comprehensive experiments to assess the capabilities of models based on PBLBench with 15 state-of-the-art MLLMs/LLMs.
Experimental results reveal that models still struggle to handle complex projects involving cross-modal representations.
For example, the o4-mini model achieved the highest ranking accuracy, which was only 59\%. These models exhibit significant hallucinations and instability, particularly in PBL scenarios where key information is missing, leading to inaccurate or incomplete assessments.
We highlight our key contributions below:
\vspace{-0.3em}
\begin{itemize}[leftmargin=*]
\item To the best of our knowledge, we built the first multi-modal PBL-STEM dataset, which contains over 500 high-quality student projects and various modalities. This contributes to filling the dataset gap in the STEM domain.
\item To evaluate current MLLMs' performance on the STEM-oriented PBL outcomes assessment task, we construct PBLBench that rigorously evaluates projects with long-context inputs and joint representations of multiple modalities. We also introduce the AHP, which employs expert-driven pairwise comparisons to derive evaluation criteria.
\item  We provide a detailed analysis of performance in existing state-of-the-art MLLMs, highlighting key performance, hallucinations, and behavioral differences under challenging multi-modal conditions. We hope PBL-STEM dataset and benchmark can bring more influence and help other researchers to develop more powerful tools for the AI4Edu research community.
\end{itemize}
\vspace{-0.3em}




%% file: sec/2_method.tex
\vspace{-0.3em}
\section{PBL-STEM Dataset and PBLBench}

\subsection{PBL-STEM Dataset}
We propose a new multimodal dataset, PBL-STEM, designed to benchmark the capability of current MLLMs in evaluating PBL outcomes with complex and diverse representations.
Unlike previous datasets, PBL-STEM focuses on complex in-context scenarios, where MLLMs must comprehend the entire project holistically and draw conclusions by integrating multidimensional knowledge, which involves project-related reports, images, slides, videos, and code.
The PBL-STEM dataset comprises a total of 500 PBL outcomes, covering the following different modalities:
\textbf{Text}: student-submitted project reports, which serve as the primary basis for PBL assessment. Furthermore, the introduce of project background is also included in the PBL-STEM dataset.
\textbf{Image}: which include circuit design diagrams, PCBs, chemical molecular structures, or diagrams of neural networks. Additionally, images in PBL also include slides submitted by students.
\textbf{Code}: core code related to target project, involving programming languages such as C, C++, and Python.
\textbf{Video}: a communication medium in PBL, enabling students to provide a more comprehensive introduction to their projects.

The PBL-STEM dataset is structured with multiple key subjects:
\textbf{Science}: chemical experiment design, which refers to adapting scientific research findings into basic or comprehensive experiments suitable for undergraduate teaching needs. In the science projects, students record videos and write reports to build a systematic project. Considering modality compatibility, videos are transformed into summaries, thus ensuring their acceptance by all models.
\textbf{Technology}: artificial intelligence applications leverage image classification algorithms to control robotic arm grabbing, achieving a combination of computer vision, artificial intelligence, and automation engineering technologies. The project materials encompass a report, images, code, and slides.
\textbf{Engineering}: embedded development involves the development and implementation of complete embedded solutions, including hardware design, software programming, and system testing. For the code of project, we manually select the core code from the project and drop those library codes which has been learned by MLLMs. The circuit design diagrams are transferred to the format acceptable to MLLMs for testing.
\textbf{Mathematics}: mathematical modeling competition, which tests students' abilities to model and solve mathematical problems. For projects in math, students need to present the solutions with code and text. To explore the gap in project evaluation capabilities across different languages using MLLMs, the PBL-STEM dataset includes mathematical modeling competitions in both Chinese and English.

\begin{table}[ht]
	\centering
   \vspace{-3mm}
	\caption{Comparison between PBLBench and the current STEM Benchmark, which involves the comparison of modalities, disciplines, and types of answers.}
	\setlength{\tabcolsep}{1.25pt} 
    \begin{tabular}{c|ccc|cccc|cccc}
        \bottomrule[1.2pt]
        Benchmark                         & Image  & Code  & Video & Sci & Tech & Eng & Math &  AnswerType & H/GEvaluation & \#Models\\
        \bottomrule[1.2pt]
        MMLU~\cite{hendrycks2024measuring} & \xmark & \xmark & \xmark  & \cmark & \cmark & \cmark & \cmark & Open/MC & Human\&GPT & 5\\
        SCIENCEQA~\cite{lu2022learn}  & \cmark & \xmark & \xmark  & \cmark & \xmark & \cmark & \xmark &  MC & - & 12 \\
        STEM~\cite{shen2024measuring}  & \cmark & \xmark & \xmark  & \cmark & \cmark & \cmark & \cmark & MC & - & 8 \\
        DYNAMATH~\cite{zoudynamath}  & \cmark & \cmark & \xmark & \cmark & \xmark & \xmark & \cmark & MC & - & 8 \\
        MMMU~\cite{yue2023mmmu}  & \cmark & \xmark & \cmark & \xmark & \cmark & \cmark & \cmark & MC & - & 14 \\
        NOVELQA~\cite{wang2024novelqa} & \xmark & \xmark & \xmark & \cmark & \cmark & \cmark & \cmark & MC & - & 7 \\
        \hline
        PBLBench(Ours)  & \cmark & \cmark & \cmark & \cmark & \cmark & \cmark & \cmark & \textbf{Free-form} & \textbf{Human} & \textbf{15}\\
        \bottomrule[1.2pt]
	\end{tabular}
    \label{tab_datset}
\end{table}
\textbf{Comparison with Existing Datasets:} The PBL-STEM dataset is the first multimodal STEM dataset for project-based learning, which encompasses multiple modalities including text, images, code, and video. As Table \ref{tab_datset} illustrates, the PBL-STEM dataset, featuring more comprehensive subject coverage and richer modalities than previous datasets, enhances the depth of evaluation for MLLM performance. 
In addition, we also compared the types of answers, and our evaluation involved open-ended answer formats.
%
\vspace{-0.3em}

\begin{figure}[!t]
    \centering
    \includegraphics[width=\linewidth]{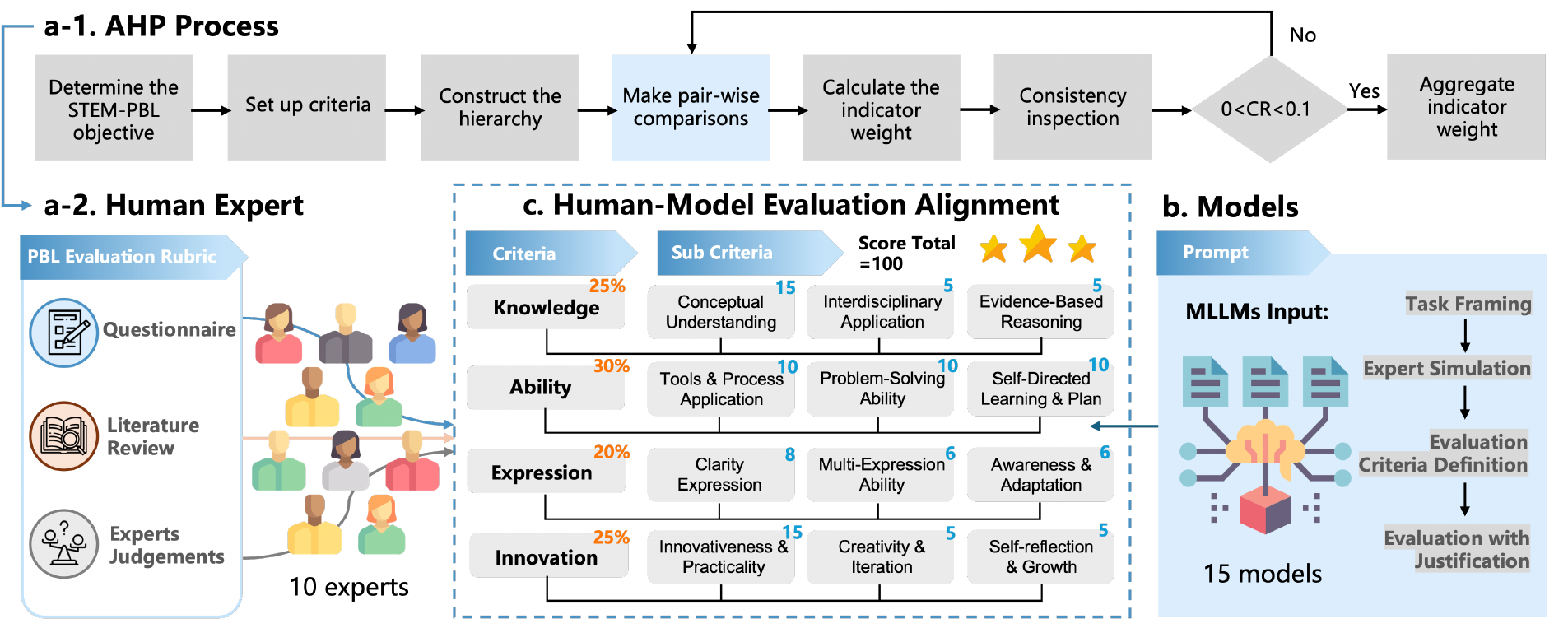}
    \vspace{-5mm}
    \caption{The pipeline of PBLBench includes the construction of evaluation criteria, scoring by human experts, and model scoring processes.}
    \vspace{-6mm}
    \label{fig:pipeline}
\end{figure}

\subsection{PBLBench} \label{PBLBench}
\vspace{-0.3em}
\textbf{Task Definition:}
In project-based learning (PBL), teachers need to score and rank the outcomes submitted by students.
Therefore, we formalize the task definition of PBLBench to leverage MLLM or LLM to automatically score projects and rank them according to their scores. Consider a PBL outcomes, which includes multiple students:
\begin{equation}
{(\text{S}_i; \text{R}_i)}_{i=0}^n ={\text{M}(\text{P}, \text{x}_i)}_{i=0}^n; \; \forall \text{x}\in \text{PBL-STEM},
\end{equation}
where $\text{S}$ denotes the score, $\text{R}$ represents the ranking, $\text{M}$ stands for the model, $\text{P}$ indicates prompts written by humans, which involve evaluation criteria for the project.

\textbf{Evaluation Criteria:}
Considering the complexity of multimodal project-based learning evaluation, PBLBench needs to individually understand the reports, code, images, videos, and other materials submitted by students.
To construct a structured and weighted evaluation framework for evaluating multimodal PBL outcomes, the Analytic Hierarchy Process (AHP) was employed~\cite{ewis2023assessment}, as shown in Figure \ref{fig:pipeline}. 
Drawing on relevant literature~\cite{gao12024check} and domain analysis\footnote{\url{https://my.pblworks.org/resource/document/project_design_rubric}}\textsuperscript{,}\footnote{\url{https://my.pblworks.org/resource/document/pbl_essential_elements_checklist}}, the evaluation framework was organized into four primary dimensions—knowledge, skills, expression, and innovation \& reflection—further subdivided into twelve secondary indicators: 
(\textbf{i}) Knowledge dimension, which involves understanding of concepts, interdisciplinary applications, and evidence comprehension skills. 
(\textbf{ii}) Skills dimension, which includes the application of tools, problem-solving abilities, and self-regulation and planning.
(\textbf{iii}) Expression dimension, which involves clarity of information, multimodal expression capabilities, and audience awareness and adaptability.
(\textbf{iv}) Innovation \& Reflection dimension, which includes innovation and practicality, innovation development and iteration, and self-reflection and growth.
We leverage manually crafted prompts to guide models to assess the quality of the project and to score it based on the aforementioned four dimensions and their twelve secondary indicators.

For the calculation of weights for each secondary indicator, a panel of ten experts with experience in STEM education and PBL pedagogy participated in a pairwise comparison process using structured AHP questionnaires.
The resulting judgment matrices were tested for consistency, with all Consistency Ratios (CR) maintained below 0.1 to ensure acceptable logical coherence.
Individual expert matrices were then aggregated using the geometric mean method to derive the final weight for each indicator.
This process enabled the development of a transparent, expert-validated, and quantitatively grounded rubric for evaluating PBL outcomes.

%% file: sec/3_experiment.tex
\vspace{-0.3em}
\section{Experiment}
\vspace{-0.3em}
\subsection{Experiment Details}
\textbf{Model Selection:}
We benchmark a range of current state-of-the-art foundation models on PBL-STEM, covering multimodal large language models (MLLMs), large language models (LLMs), and reasoning models. Our evaluation includes ten recent closed-source models~(o4-mini~\cite{openai-o3-o4-mini-system-card-2025}, GPT-4.1~\cite{openai_gpt4_1}, GPT-4.1-mini~\cite{openai_gpt4_1}, GPT-4o~\cite{openai_gpt4o_mini}, GPT-4o-mini~\cite{openai_gpt4o_mini}, Gemini-2.0-flash~\cite{pichai2024introducing}, Gemini-2.5-flash~\cite{google-gemini2.5-flash-preview-2025}, Claude-3.7~\cite{Claude3S}, Phi-4~\cite{abdin2024phi}, and Grok-3~\cite{xai-grok3-2024}) accessed via their respective APIs. Additionally, we deploy five leading open-source models~(DeepSeek-V3~\cite{liu2024deepseek}, DeepSeek-R1~\cite{guo2025deepseek}, LLaMA-4~\cite{meta2025llama}, Qwen-2.5~\cite{yang2024qwen2} and Qwen-3~\cite{qwen3-2025}) as well as one advanced multi-modal model~(LLaVA-1.6~\cite{liu2023improved}), running all local models on a cluster of 4 NVIDIA A6000 GPUs.
It should be noted that the Qwen-2.5, Qwen-3, LLaMA-4, DeepSeek-R1, and Grok-3 models do not possess multimodal processing capabilities.
Consequently, in the experiments involving the aforementioned models, our inputs are converted to the text modality only.
To further explore the impact of reasoning capabilities, we also evaluate two dedicated reasoning models (Gemini-2.5-flash with thinking~\cite{google-gemini2.5-flash-preview-2025} and DeepSeek-R1~\citep{guo2025deepseek}) to assess the benefits of deep thinking on PBLBench tasks. Additionally, we evaluate the reasoning abilities of the GPT-4o~\cite{openai_gpt4o_mini} model with Chain-of-Thought.

\textbf{Metrics:}
For the evaluation metrics, we use average scores (\textbf{Ave}) along with standard deviation (\textbf{Std}), and compare these to human scores based on predefined evaluation criteria. 
Additionally, we rank the projects based on human scores and report the accuracy of the model's rankings  (\textbf{Acc}).
To avoid randomness and hallucination in model outputs, each project is assessed five times in our experiments, filtering out the highest and lowest scores to calculate the average and standard deviation.

\vspace{-0.3em}
\subsection{Experiment Results}
\vspace{-0.3em}
To verify the potential of MLLMs in evaluation PBL outcomes, we are conducting detailed experiments on the PBL-STEM dataset, which includes various disciplines.
The results of the experiments are shown in Table \ref{tab_main}, from which the following conclusions can be drawn:

\begin{table*}[ht]
\centering
\vspace{-0.5\intextsep}
\caption{Results of the \textbf{average scores with standard deviation} and \textbf{ranking accuracy} under different disciplinary, MLLM, and LLM settings.}
\vspace{-0.5\intextsep}
\setlength{\tabcolsep}{0.78mm}        
{{
\begin{tabular}{c|ccc|ccc|ccc|ccc}
    \toprule[1.5pt]
    \multirow{2}{*}{\textbf{Model}} & 
    \multicolumn{3}{c|}{\textbf{Science}} & 
    \multicolumn{3}{c|}{\textbf{Technology}} &  
    \multicolumn{3}{c|}{\textbf{Engineering}} &  
    \multicolumn{3}{c}{\textbf{Mathematics}} \\
    
\cmidrule(rl){2-4}\cmidrule(rl){5-7} \cmidrule(rl){8-10} \cmidrule(rl){11-13} 
& {Ave} & {Std} & {Acc} & {Ave} & {Std} & {Acc} & {Ave} & {Std} & {Acc} & {Ave} & {Std} & {Acc} \\
\hline
LLaVA-1.6                 &43.56	&14.30	&21.05      &47.47	&25.29	&15.78                            &22.54	&11.54	&22.72      &17.11	&8.79	&24.82\\
Qwen-2.5                  &88.53	&1.90	&15.78      &\textbf{84.71}	&2.21	&42.10        &79.39	&\underline{1.53}	&18.18      &71.47	&4.55	&27.53 \\
Qwen-3                    &87.50	&2.61	&31.57      &66.29   &6.32	&26.31                            &61.73	&4.14	&22.72      &68.90	&5.12	&27.94 \\
LLaMA-4                   &81.32	&1.22	&26.31      &75.78	& \underline{1.61}	&26.31                &78.78	&1.90	&27.27      &76.26	& \textbf{1.47}	&26.82 \\
DeepSeek-V3               &96.66	&3.42	&31.57      &73.57	&13.84	&31.57                            &86.20	&3.97	&36.36      &71.72	&6.42	& 34.66 \\
DeepSeek-R1               &69.06	&17.95	&26.31      &69.85	&13.72	&26.31                            &61.43	&14.28	&36.36      &74.91	&14.2	& \underline{35.38} \\
\cmidrule(rl){1-13}
Phi-4-mul                 &67.35	&4.31	&26.31      &\underline{83.99}	&6.62	& \underline{47.36}      &62.92	&10.16	&22.72      & \textbf{81.98}	&11.56	&26.10 \\
Grok-3                    &85.63	&4.89	&10.52      &72.14	&4.69	&26.31                            &80.34	&1.99	&22.72      &76.54	& 1.91	&27.53 \\
Gemini-2.5-flash          &93.83	&1.84	&36.84      &69.29	&4.66	&36.84                &79.48	&2.54	&\textbf{50.0}      &73.63	&6.27	& 33.94 \\
Claude-3.7                &93.82	& \textbf{0.93}	&31.57      &78.33	& \textbf{1.81}	&31.57            &80.32	&\textbf{1.51}	&\underline{45.45}      &68.55	&2.13	&28.82 \\
o4-mini                   &63.75	&2.72	&\textbf{47.38}      &79.96   &2.32	&\textbf{59.0}                             &62.59	&2.12	&42.10      &89.52	&1.52	&\textbf{36.84} \\
GPT-4o-mini               &85.66	&2.98	&31.57      &76.50	& 2.84	&26.31                &\underline{82.01}	&3.16	&31.81      &79.06	&2.90	&20.97 \\
GPT-4o                    & \underline{92.99}	&1.98	&26.31      & 83.45	&2.89	& \underline{47.36}      &89.48	&1.97	&40.90      & \underline{79.10}	&2.54	&32.10 \\
GPT-4.1-mini              & \textbf{89.99}	&2.18	&  \underline{42.10} &74.82	&3.05	&21.05        &\textbf{83.79}	&1.79	&\underline{45.45}      &71.47	&3.63	& 33.38 \\
GPT-4.1                   &94.31	& \underline{1.15}	&31.57      &77.26	&3.06	&21.05                &85.79	&2.47	&36.36      &74.19	& \underline{1.77}	&32.10 \\
\hline
\rowcolor{aliceblue!60} Human  &90.68	&1.08	&-      &84.94	&1.21	&-     &83.27	&1.90	&-      &84.83	&1.27	    &- \\
\toprule[1.5pt]
\end{tabular}}}
\vspace{-0.5\intextsep}
\label{tab_main}
\end{table*}
\textbf{Scoring Consistency:}
(\textbf{i}) Ideally, the scoring of feasible MLLMs should closely approximate that achieved by human assessors. By observing Table \ref{tab_main}, we find that the scoring of GPT-series models is more applicable in the disciplines of science and engineering.
In the discipline of science, the GPT-4.1-mini model achieves a score of 89.99, closely matching the human benchmark with a mere difference of 0.69. In technology, the Qwen-2.5 model scores 84.71, just 0.23 short of the human evaluation. In engineering, the GPT-4.1-mini model scores 83.79, narrowly trailing the human score by 0.52. In mathematics, the Phi-4-mul model scores 81.98, which is 2.85 less than the human score.
(\textbf{ii}) In all disciplines, the model demonstrates a superior evaluative capacity in the domain of science, evidenced by an average deviation from human assessments of 9.69. In comparison, the average deviations for the domains of technology and engineering are 11.13 and 11.14, respectively, whereas the domain of mathematics exhibits a higher deviation of 14.48.
(\textbf{iii}) Compared to open-source models, the evaluations of closed-source models more closely approximate human scoring. Among the four disciplines, the closed-source models GPT-4.1-mini and Phi-4-mul achieve the optimal scores in the fields of science, engineering, and mathematics.
(\textbf{iv}) Compared to open-source models, closed-source models have a lower standard deviation. It is worth noting that the Claude-3.7 model exhibits the lowest standard deviation in the disciplines of science, technology, and engineering. This indicates that the Claude-3.7 model possesses greater stability and fewer hallucination issues compared to other models.
(\textbf{v}) Upon closer examination of the performance of different models, the DeepSeek-R1 model generally scores lower than the DeepSeek-V3 across most disciplines. The GPT-4o or GPT-4.1 models score higher than their corresponding mini versions and exhibit relatively more stable standard deviations. For instance, in all disciplines, the average score of the GPT-4o model surpasses that of its corresponding mini version by a margin of 5.44. Moreover, the performance of the latest Qwen-3 model is not as optimal as that of the Qwen-2.5 model, which exhibits lower scores and higher standard deviations.

\textbf{Ranking Accuracy:}
(\textbf{i}) In the verification of ranking accuracy, we observed that the o4-mini model achieved the best ranking in the subjects of science, technology, and mathematics, despite having lower scores.
(\textbf{ii}) Compared to open-source models, closed-source models achieve higher accuracy in rankings. For instance, in the discipline of technology, the o4-mini model achieves a ranking accuracy of 59\%. 
(\textbf{iii}) In the GPT-4o model, the mini version ranks lower than the full version in most disciplines. However, the GPT-4.1-mini model outperforms GPT-4.1. Additionally, the Gemini-2.5-flash model achieves the highest ranking accuracy in the discipline of engineering.
From the results above, we observe that although the model's average scores are close to those of humans, it demonstrates lower ranking accuracy, which indicates that the model is not effectively evaluating PBL outcomes.


\vspace{-0.3em}
\subsection{Ablation Experiment and Discussion}
\begin{figure*}[h]
\vspace{-0.75\intextsep}
  \centering
  \captionsetup[subfloat]{font=scriptsize}
  \subfloat[Different languages]{\includegraphics[width=0.49\textwidth]{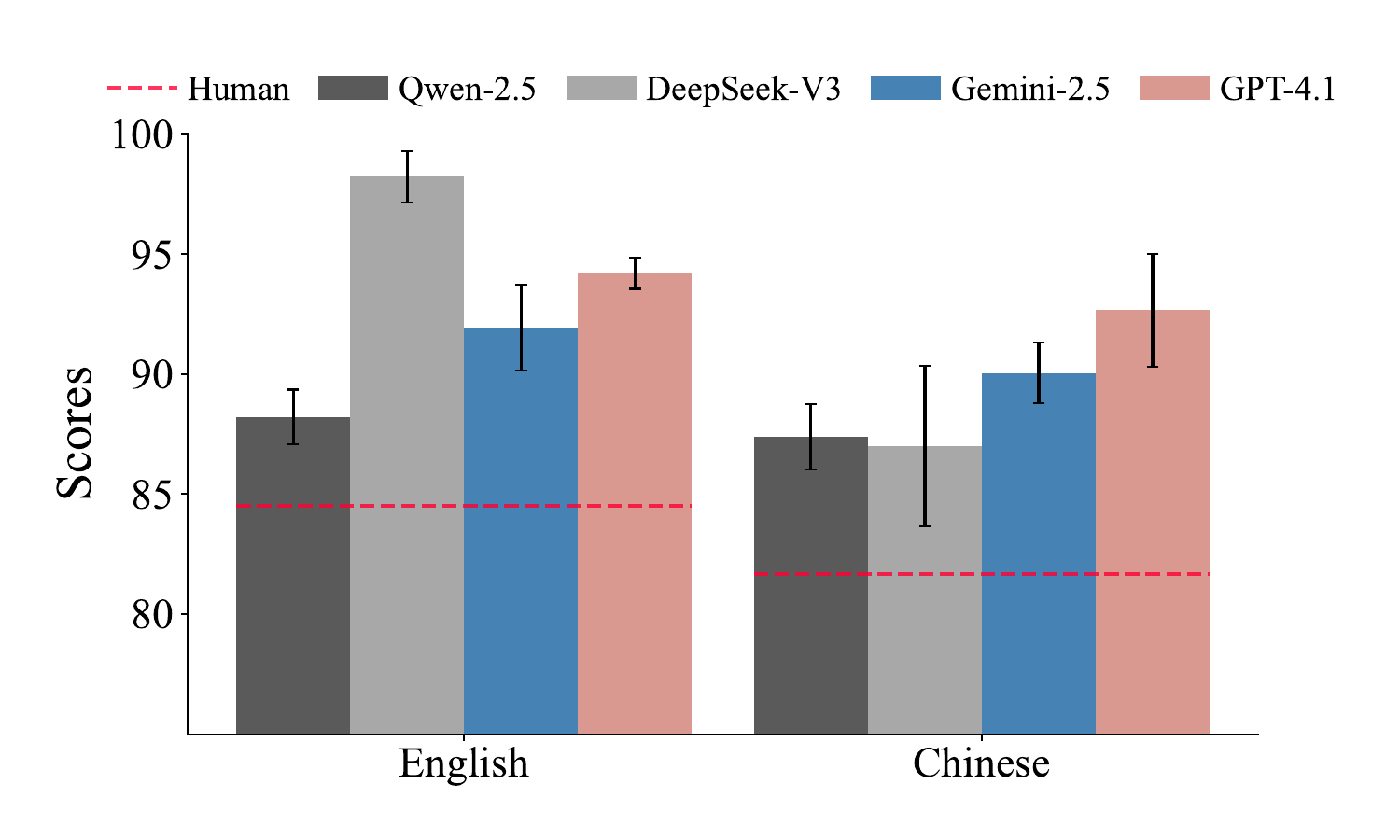}\label{fig:4.1}}
  \subfloat[Different modalities]{\includegraphics[width=0.49\textwidth]{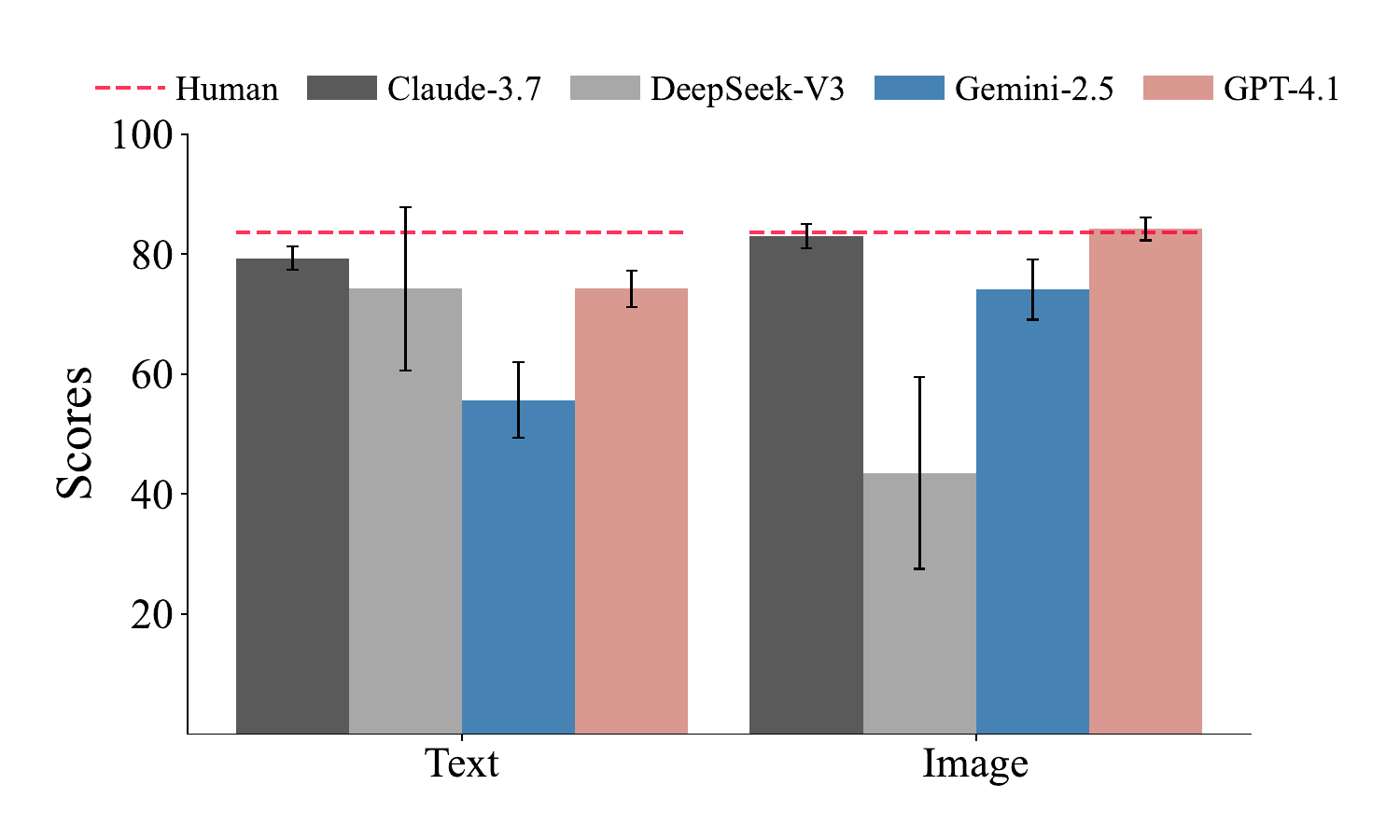}\label{fig:4.2}}
\vspace{-0.5\intextsep}
\caption{The results compare the performance of \textbf{different languages} and \textbf{different modalities}.}
\vspace{-0.5\intextsep}
\label{figure:4} 
\end{figure*}

\vspace{-0.3em}
\textbf{Different Languages Evaluation:}
To verify the models' ability to evaluate projects in different languages, as shown in Figure \ref{fig:4.1}, we conducted a scoring comparison between settings in Chinese and English.
In the evaluation of Chinese projects, the Qwen-2.5 model demonstrates superior evaluative capabilities compared to the Gemini-2.5 and GPT-4.1 models, aligning with the descriptions provided by the Qwen-2.5 technical report~\cite{yang2024qwen2}.
Furthermore, we observed that the GPT-4.1 model exhibits a lower standard deviation compared to other models in English settings, indicating greater stability.
Lastly, in the evaluation of Chinese projects, the standard deviation is often higher compared to the assessment of English projects, indicating that the models' understanding of Chinese evaluation tasks is less stable, potentially leading to issues of inconsistency.

\textbf{Different Modalities Evaluation:}
Figure \ref{fig:4.2} illustrates the evaluative capabilities of the model for projects based on different modalities.
This involves the transformation of project reports into textual format or the incorporation of images as the sole input.
We observed that the evaluative capabilities of the Claude-3.7 model, which uses text and images as inputs, are superior to those of the other three models and demonstrate remarkable stability.
Secondly, the evaluative capabilities of the DeepSeek-V3 model for target text are significantly better than for images, which more closely approximate human scoring.
Furthermore, the Gemini-2.5 and GPT-4.1 models demonstrate superior evaluative capabilities with images compared to text, and they exhibit smaller standard deviations.

\begin{figure*}[ht]
\vspace{-1mm}
  \centering
  \captionsetup[subfloat]{font=scriptsize}
  \subfloat[Different model sizes]{\includegraphics[width=2.75in]{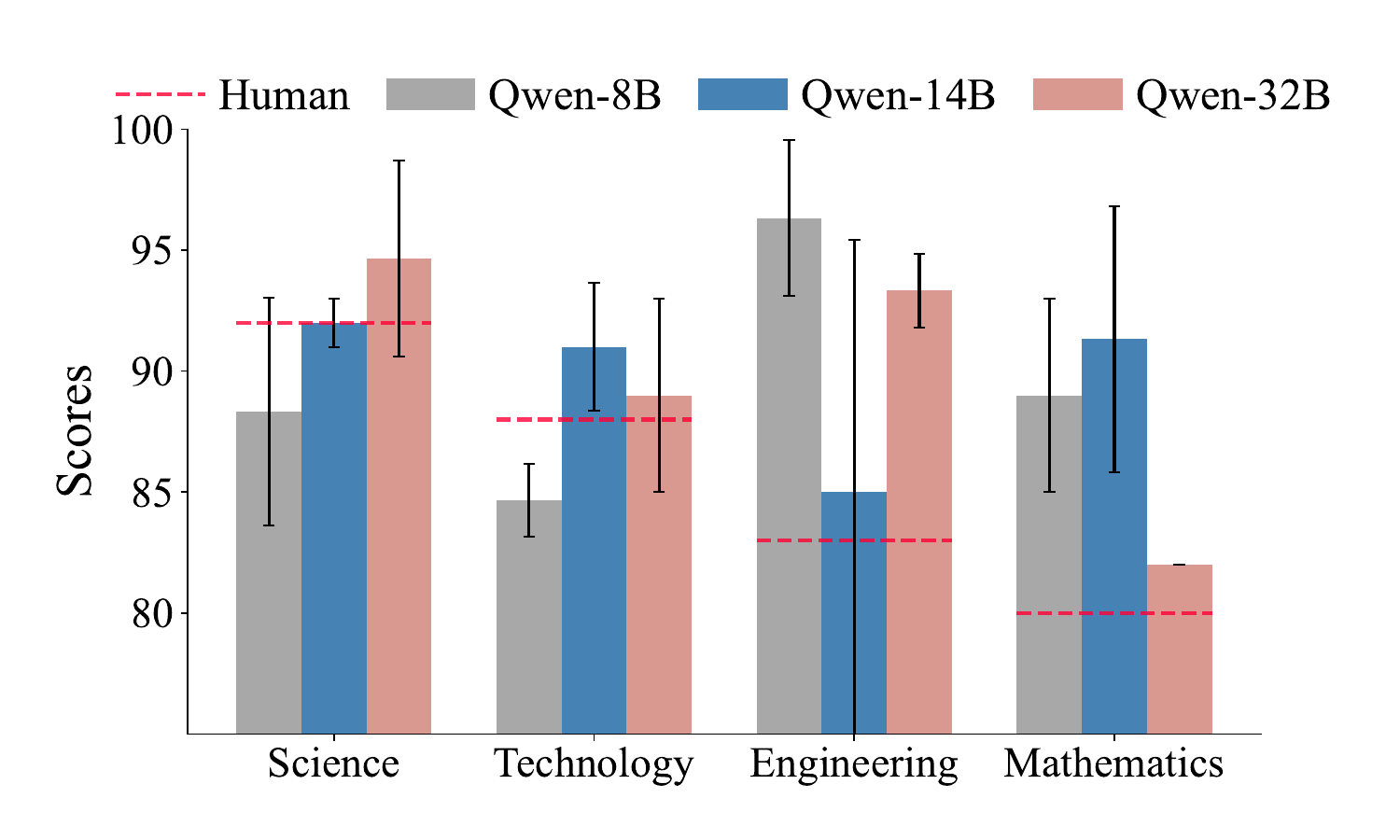}\label{fig:3.1}}
  \subfloat[Different materials]{\includegraphics[width=2.75in]{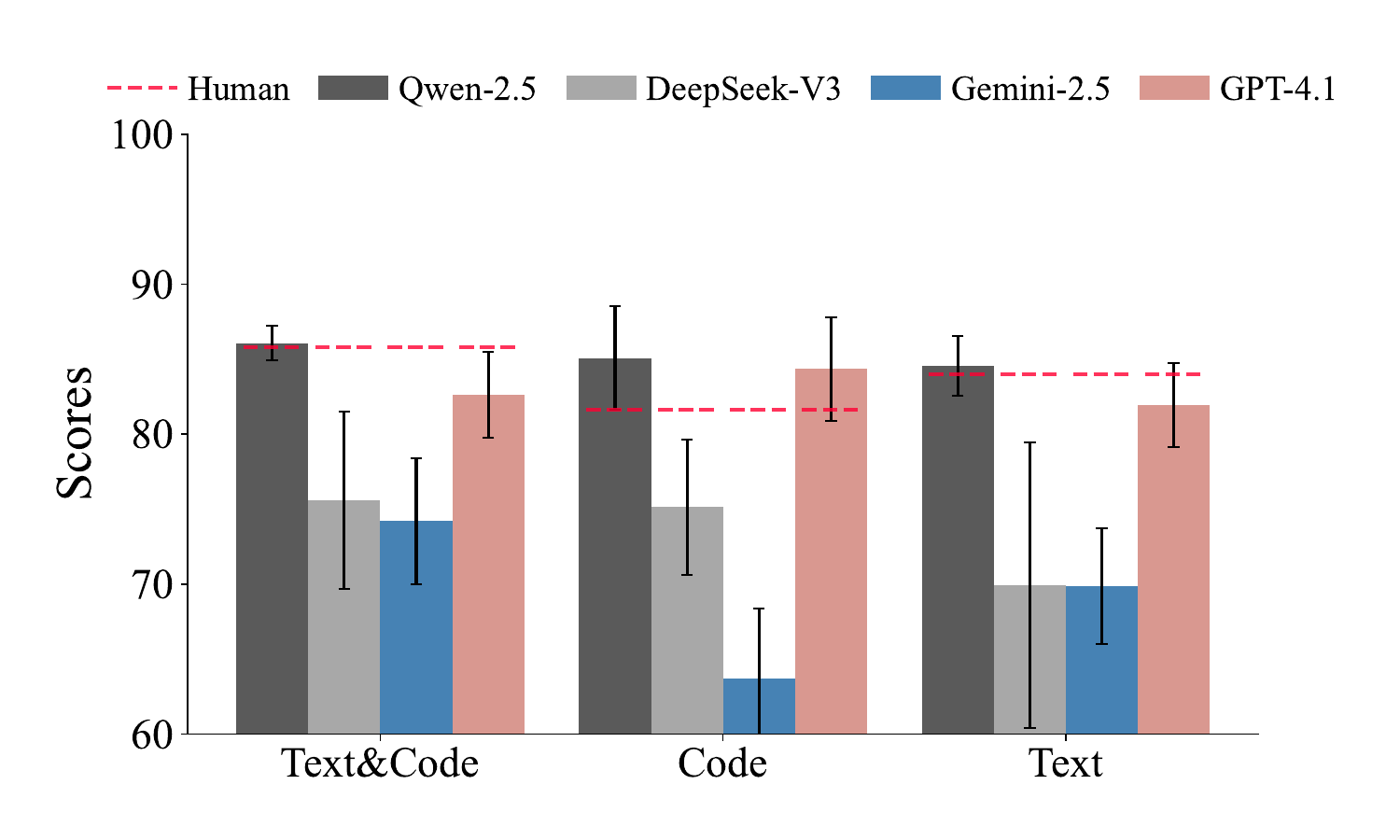}\label{fig:3.2}}
\vspace{-0.5\intextsep}
\caption{The results compare the performance of \textbf{different model sizes} and \textbf{various materials}.}
\vspace{-0.5\intextsep}
\label{figure:3} 
\end{figure*}

\textbf{Model Size Evaluation:}
Figure \ref{fig:3.1} demonstrates the impact of varying model sizes on the model's evaluative capabilities.
Firstly, we observe that with increasing model size, the scoring fails to remain aligned with human scores.
Secondly, the Qwen3-14B model's scoring is more closely aligned with human evaluations in the disciplines of science, technology, and engineering, but it exhibits a significant standard deviation.
Additionally, in the discipline of mathematics, as the model size increases, the Qwen3-32B model's ability to understand the project improves, leading to scores that are closer to human evaluations and a standard deviation approaching zero.
In summary, increasing the size of the model does not consistently lead to improvements in assessment performance.

\textbf{Different Materials Evaluation:}
We verify the evaluative capabilities of the model for projects based on different materials, as shown in Figure \ref{fig:3.2}.
It is not difficult to observe that the Qwen-2.5 model performs excellently, more closely aligning with human scoring compared to other models, and exhibits a smaller standard deviation.
Furthermore, compared to the Gemini-2.5 and DeepSeek-V3 models, the Qwen-2.5 and GPT-4.1 models offer superior evaluation capabilities for code. The same advantage is observed in the text modality.
Finally, although providing detailed project materials such as text and code enhances the model's understanding of the project, the Gemini-2.5 model still fails to align with human scores.

\begin{figure*}[h]
\vspace{-0.5\intextsep}
  \centering
  \captionsetup[subfloat]{font=scriptsize}
  \subfloat[Conversion for video]{\includegraphics[width=2.75in]{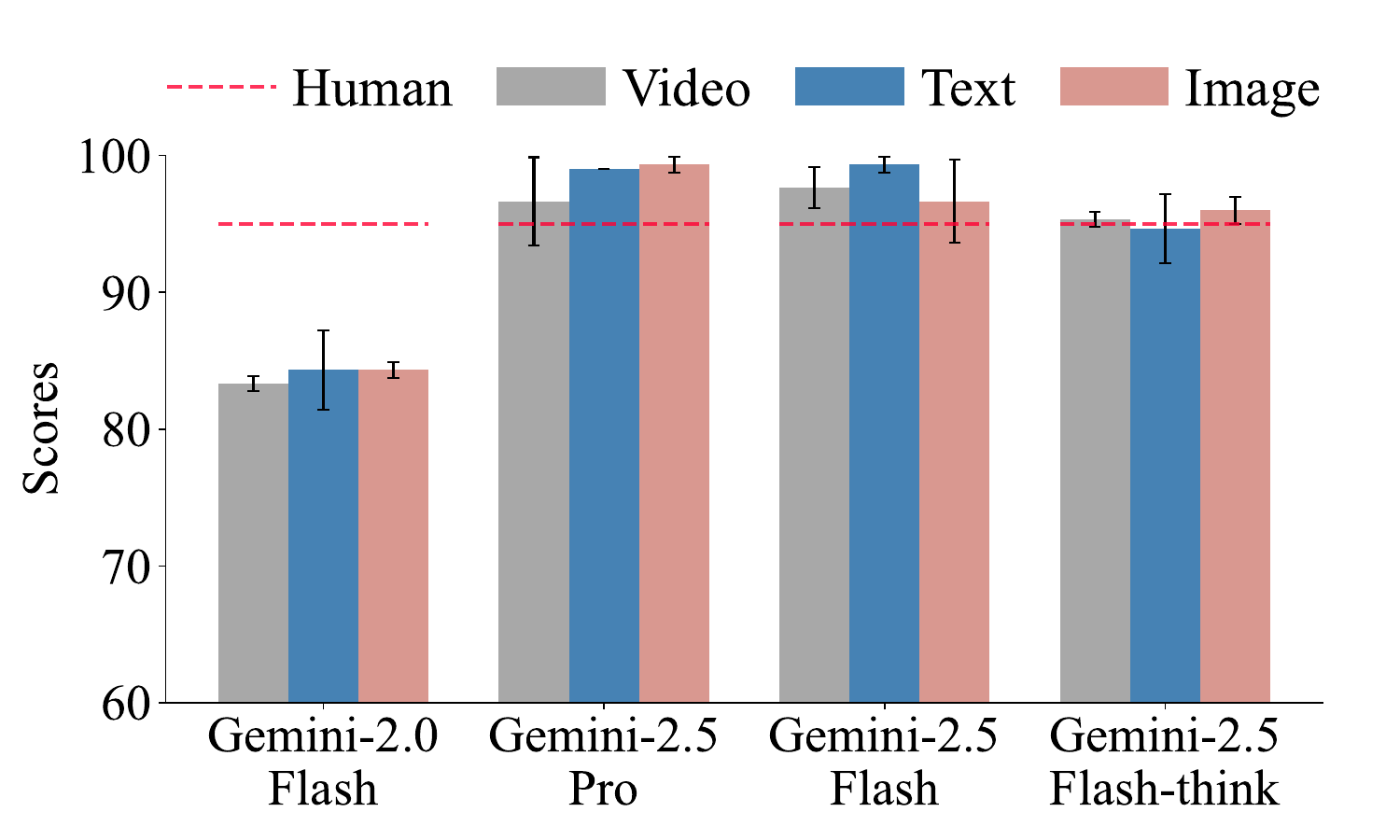}\label{fig:5.1}}
  \subfloat[Different input lengths]{\includegraphics[width=2.75in]{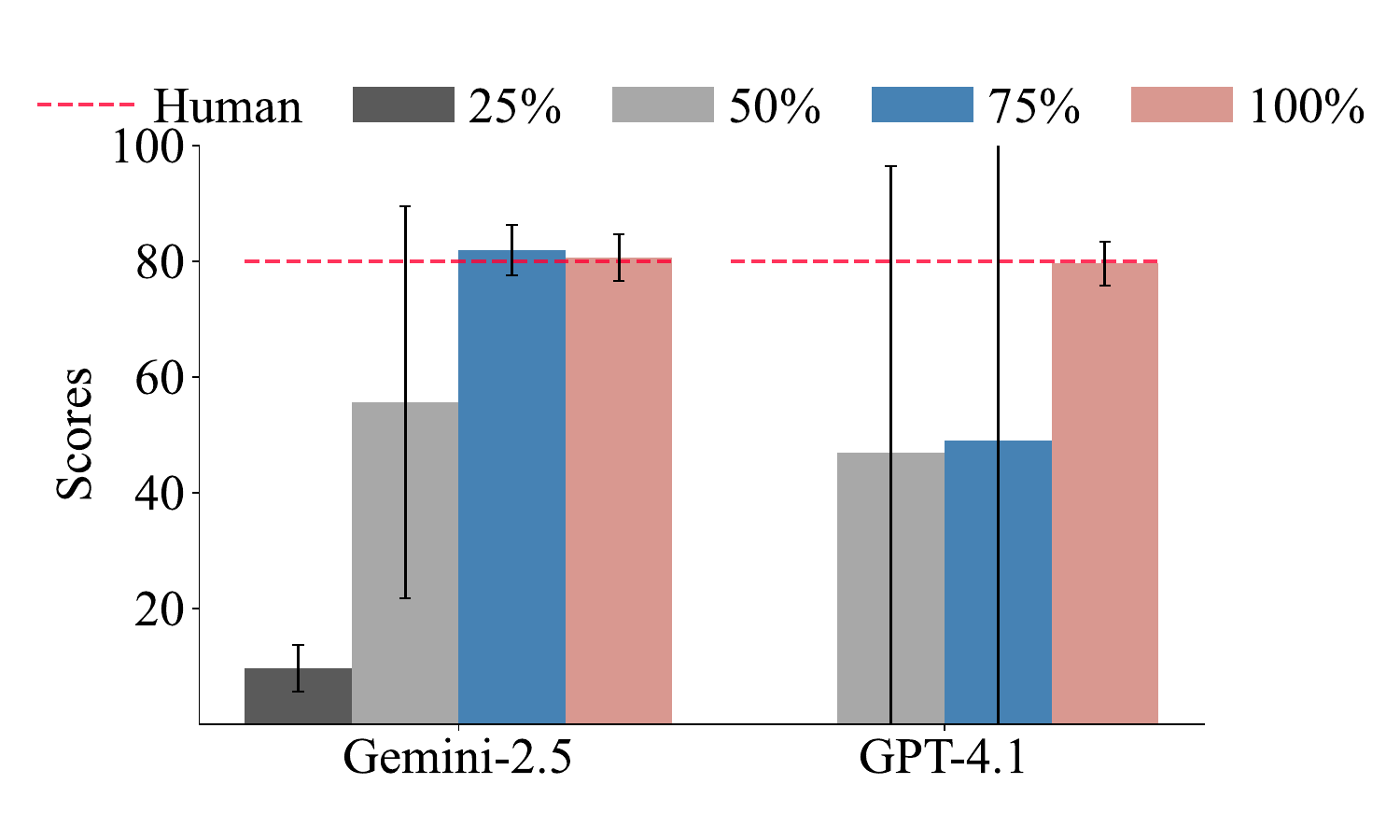}\label{fig:5.2}}
\vspace{-0.5\intextsep}
\caption{Performance comparison of the \textbf{type of video} and the \textbf{different input lengths}.}
\vspace{-0.75\intextsep}
\label{figure:5} 
\end{figure*}

\textbf{Evaluation of Video Processing Approaches:}
Considering that models like Claude-3.7 lack effective video processing capabilities, we converted videos into summaries for our experiments.
To explore the impact of different video formats on project assessment, we compare the evaluation performance of three formats—video, video to text, and video to image—in the Gemini model.
In the video-to-text setting, we convert the video content into textual summaries, and in the video-to-image setting, we transform the video into images.
As shown in Figure \ref{fig:5.1}, the scores from the Gemini-2.5-flash with thinking model closely approximate human scores across different modalities.
In the Gemini-2.5-Pro model, the evaluation performance on original videos is superior to that of text and images, which may be due to the loss of some video information during the conversion process.

\textbf{Completeness Evaluation:}
To further assess the model's ability to understand the completeness of the PBL project, we truncate the project reports. 
As shown in Figure \ref{fig:5.2}, when the input document contains only a quarter of the content, the model scores lower. For instance, the GPT-4.1 model's score is 0.
As the completeness of the project reports increases, the GPT-4.1 model's performance improves, but the standard deviation significantly decreases.
However, despite only having access to three-quarters of the project reports, the Gemini-2.5 model's scoring is close to human levels, which indicates that the model may not fully understand the project reports or that it suffers from significant issues with hallucinations.

\textbf{Evaluation of Models Incorporating Thinking: }
To further explore whether deep thinking can affect the results of project evaluations, we compared the Gemini-2.5 model in thinking mode. The experimental results are shown in Table \ref{tab_think}.
In the discipline of science, it has been observed that the Gemini-2.5 model, when equipped with the thinking mode, provides evaluations that more closely approximate human-derived scores and produces rankings with enhanced precision, achieving a ranking accuracy of 47.36\%.
However, in the disciplines of technology, engineering, and mathematics, employing the thinking mode does not effectively enhance the accuracy of rankings, although the average scores are closer to those determined by human evaluators.
For example, in the discipline of technology, the gap between the average score and human scorings has narrowed by 2.39\%, but the ranking accuracy has decreased by 10.53\%.
Finally, we observed that the standard deviation of the scores in the thinking mode was smaller, which indicates that additional contemplation contributes to the consistency of the model's outputs.
\begin{table*}[ht]
\centering
\vspace{-0.5\intextsep}
\caption{Performance comparison of the \textbf{Gemini-2.5 model with thinking} for evaluating PBL.}
\vspace{-0.5\intextsep}
\setlength{\tabcolsep}{0.78mm}        
{{
\begin{tabular}{c|ccc|ccc|ccc|ccc}
    \toprule[1.5pt]
    \multirow{2}{*}{\textbf{Model}} & 
    \multicolumn{3}{c|}{\textbf{Science}} & 
    \multicolumn{3}{c|}{\textbf{Technology}} &  
    \multicolumn{3}{c|}{\textbf{Engineering}} &  
    \multicolumn{3}{c}{\textbf{Mathematics}} \\
    
\cmidrule(rl){2-4}\cmidrule(rl){5-7} \cmidrule(rl){8-10} \cmidrule(rl){11-13} 
& {Ave} & {Std} & {Acc} & {Ave} & {Std} & {Acc} & {Ave} & {Std} & {Acc} & {Ave} & {Std} & {Acc} \\
\hline
\textbf{W/o} thinking   &93.83	&1.84	&36.84  &69.29	&4.66	&36.84  &79.48	&2.54	&50.0    &73.63	&6.27	& 33.94 \\
\textbf{W}   thinking   &92.78	&1.21	&47.36  &71.68	&2.22	&26.31  &79.31	&1.81	&31.81   &73.69	&5.31	& 26.97 \\
\hline
\rowcolor{aliceblue!60} Human  &90.68	&1.08	&-      &84.94	&1.21	&-     &83.27	&1.90	&-      &84.83	&1.27	    &- \\
\toprule[1.5pt]
\end{tabular}}}
\vspace{-0.75\intextsep}
\label{tab_think}
\end{table*}

\begin{figure*}[h]
\vspace{-0.75\intextsep}
  \centering
  \captionsetup[subfloat]{font=scriptsize}
  \subfloat[Chain-of-Thought]{\includegraphics[width=2.75in]{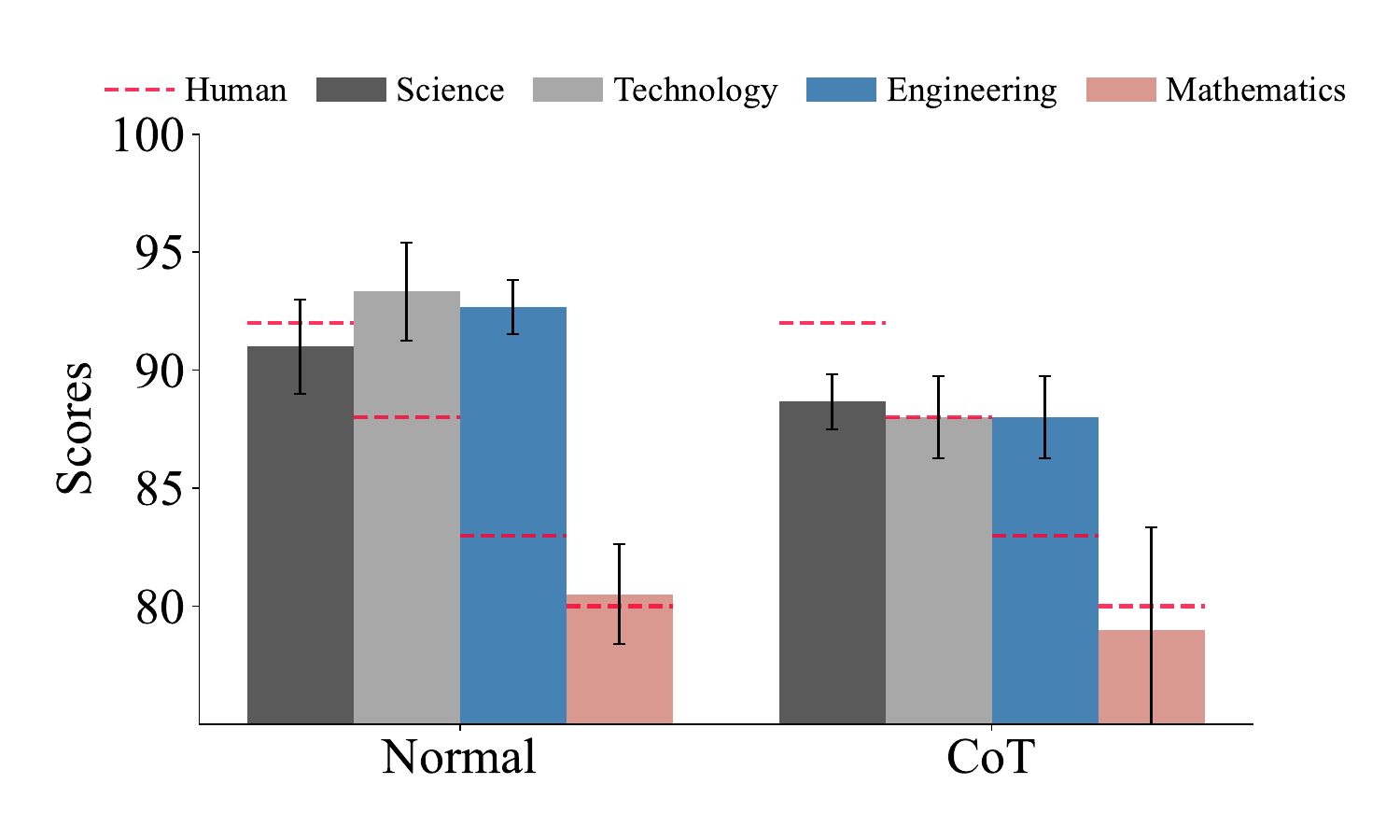}\label{fig:6.1}}
  \subfloat[Interactive]{\includegraphics[width=2.75in]{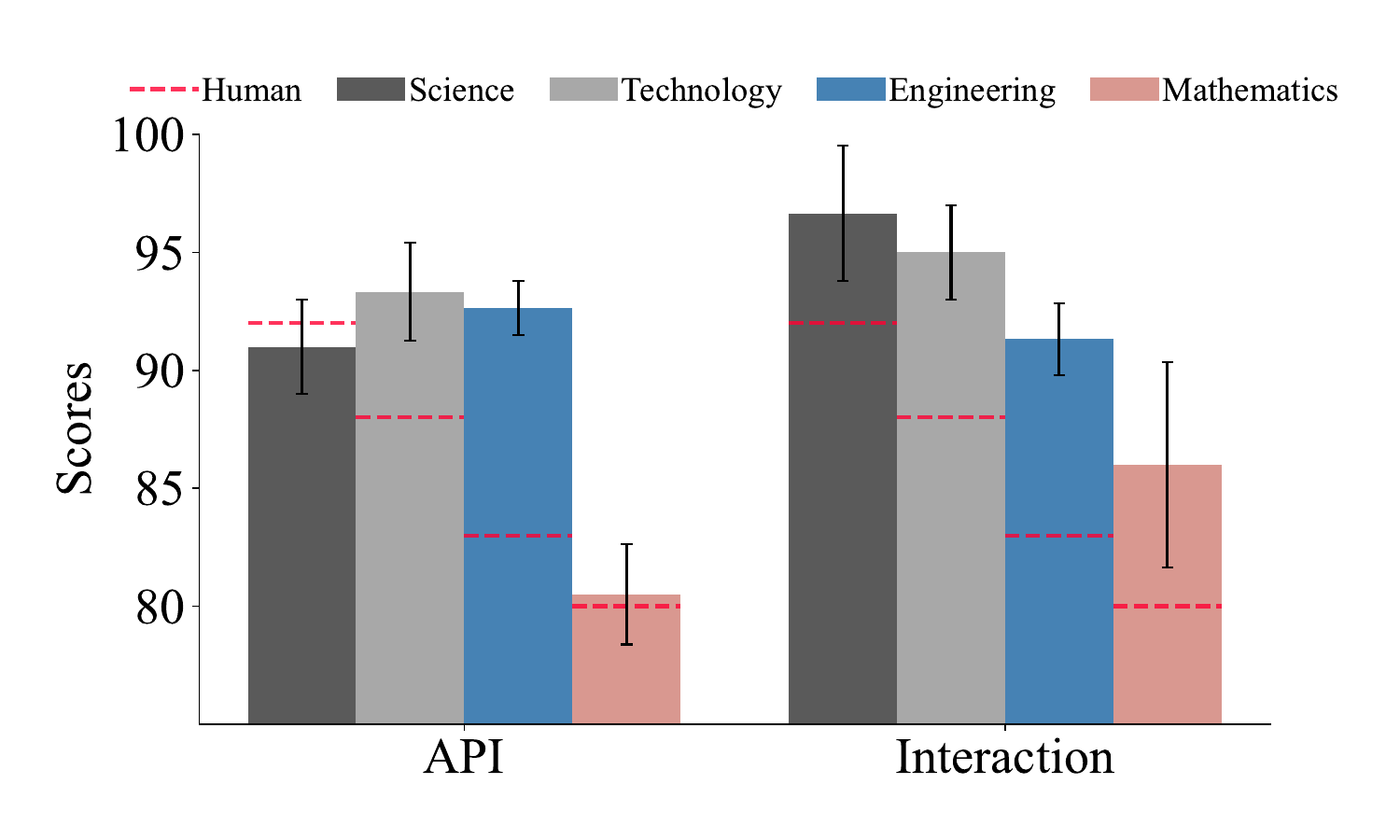}\label{fig:6.2}}
\vspace{-0.5\intextsep}
\caption{Evaluation results with \textbf{Chain-of-Thought} and comparison based on \textbf{interactive inputs}.}
\vspace{-2mm}
\label{figure:6} 
\end{figure*}

\textbf{Chain-of-Thought Evaluation:}
As shown in Figure \ref{fig:6.1}, we compare the model's ability to evaluate projects in the setting with CoT.
In evaluations with CoT, we observed that the model scores lower compared to settings without CoT in all disciplines, which suggests that rigorous thinking might lead to lower scores from the model.
In the disciplines of technology and engineering, the model with CoT scores closer to human evaluators.
Furthermore, we explored the differences between calling APIs and interactive scoring, and as shown in Figure \ref{fig:6.2}, we found that under the interactive scoring setting, the model tends to give higher scores.

\textbf{Case Study:}
In addition, we presented a case study of our PBLbench, which includes a detailed reasoning process of the MLLM in evaluating projects, as illustrated in Figure \ref{fig:case} and Table \ref{tab_case_table} in the Appendix \ref{more_experimental}.
It is evident that the model possesses multidimensional reasoning capabilities in assessing projects, including aspects such as innovation and comprehensive knowledge.
However, it is noteworthy that even when we use incomplete reports as inputs, the model still outputs seemingly reasonable reasoning processes.
This explains the reason shown in Figure \ref{fig:5.2}, which indicates that the model does not fully understand the completeness of the input report, merely outputting results without thoughtful consideration.

\begin{figure}[!t]
    \centering
    \includegraphics[width=\linewidth]{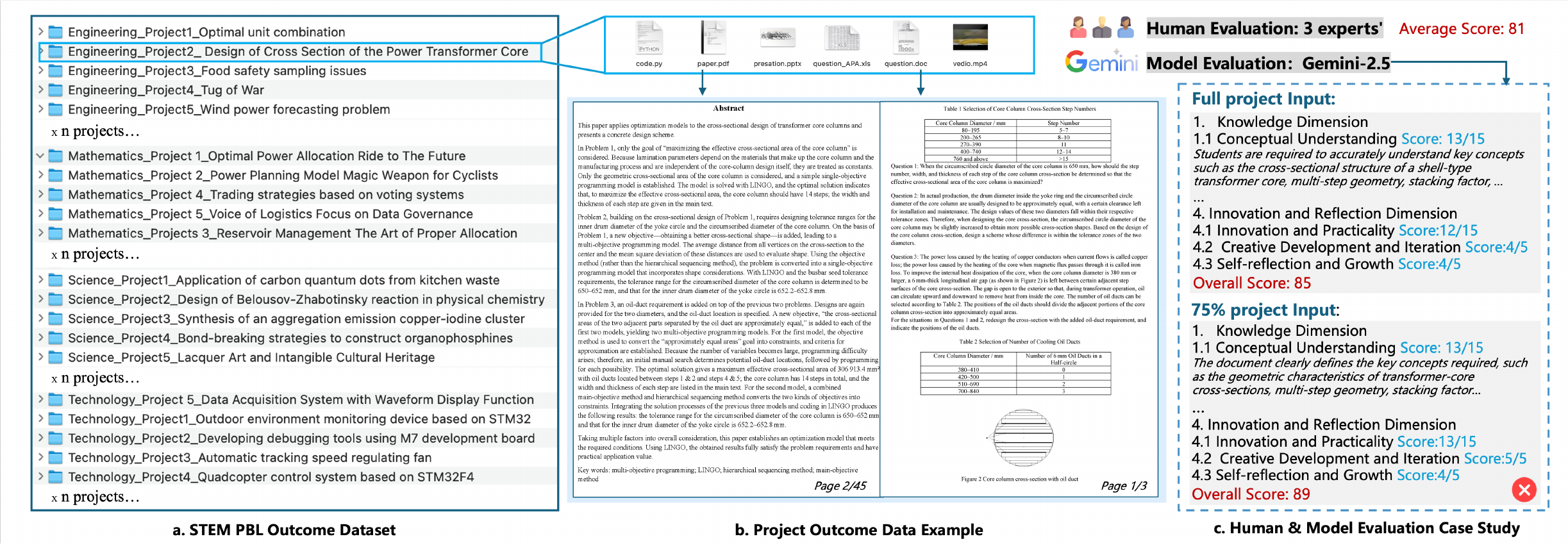}
    \vspace{-5mm}
    \caption{An example of a PBL-STEM dataset case, along with a comparison of the model's reasoning processes for different inputs.}
    \vspace{-5mm}
    \label{fig:case}
\end{figure}

\textbf{Summary and Challenges:}
Based on the results of the experiments and ablation analysis mentioned above, we have discovered that although the current state-of-the-art MLLM has the capability to score PBL outcomes, it generally exhibits significant fluctuations.
Importantly, individual models are unable to effectively assess all STEM disciplines, and they show low accuracy in rankings.
In addition, as shown in Figure \ref{fig:5.2}, the Gemini-2.5 model demonstrated a significant hallucination problem, where it continued to output high scores even for incomplete reports.
Therefore, constructing a self-verification mechanism with an agent for the MLLM to enhance scoring stability is crucial for the subsequent assessment of PBL outcomes.


%% file: sec/4_related.tex
\vspace{-1mm}
\section{Related Work}
\vspace{-1mm}
\textbf{MLLMs for Downstream Tasks:}
AI models typically tackle multiple subtasks in project-based learning scenarios, including OCR, in-context learning, video understanding, visual question answering, and multi-modal reasoning~\cite{free,zhao2024uni,xiao}. Recently, researchers have shifted toward designing automated pipelines or developing task-specific agents to manage complex tasks using MLLMs or LLMs~\cite{zhao2024unlearning,zhao2024weak}. For instance, MDAgent\cite{kim2024mdagents} employs multi-agent collaboration for medical decision-making, FinCon\cite{yu2024fincon} builds an agent system for financial decision-making through conceptual verbal reinforcement, and RestoreAgent~\cite{chen2024restoreagent} leverages MLLMs for automatic image restoration. In the context of AI4Edu, most existing approaches~\cite{chou2003redefining, viswanathan2022enhancement, xu2024eduagent} focus on integrating LLMs to support student learning, while largely overlooking the increasing workload faced by educators, especially with the expansion of higher education.

\textbf{Dataset for STEM:}
Most current STEM-related datasets and benchmarks are limited to single-discipline or simplified tasks. For example, GSM8K~\cite{cobbe2021training} and MATH~\cite{hendrycks2021measuring} focus solely on mathematics, while PIQA targets physical knowledge understanding. ARC and ScienceQA~\cite{lu2022learn} are multimodal but primarily cover general science topics. MMLU~\cite{hendrycks2024measuring} includes a broad range of 57 tasks, including STEM subjects, yet remains text-only in format. HumanEval~\cite{chen2021codex} and MBPP~\cite{austin2021program} are concentrated on programming and computer science. Though STEM~\cite{shen2024measuring} introduces a graphic and textual dataset across STEM domains, its application scenarios are limited to basic tasks like Q\&A and multiple-choice questions. Unlike previous datasets, PBL Tester is designed specifically for higher-difficulty, university-level PBL scenarios, focusing on project-based learning in higher education. The PBL-STEM component incorporates core STEM knowledge and demands cross-domain knowledge integration and multimodal representation for understanding long-term, context-rich tasks. This enables a more robust evaluation of whether current state-of-the-art models can assist or even partially replace university instructors in tasks such as grading assignments, highlighting their potential application in real educational settings.


%% file: sec/5_conclusion.tex
\vspace{-1mm}
\section{Conclusion}
\vspace{-1mm}
To evaluate the capabilities of multimodal large language models (MLLMs) in Project-Based Learning (PBL) outcomes, we introduce PBLBench, the inaugural assessment framework specifically designed for STEM-based PBL outcomes, incorporating multimodal scenarios with text, images, code, and video.
Furthermore, we construct the first multi-modal PBL-STEM dataset, which includes PBL outcomes related to STEM disciplines.
Finally, the Analytic Hierarchy Process (AHP) was introduced to construct the structured and weighted evaluation criteria.
PBLTester evaluated several state-of-the-art MLLMs, including Gemini-2.5, DeepSeek-V3, and the GPT-4o model.
The experimental results show that although the foundation models exhibit superior scoring performance, they maintain low ranking accuracy. More importantly, the models demonstrate significant instability and hallucinations.
In our future work, we will continue to enhance PBLBench to assess MLLMs. This includes evaluating the models' capabilities in multiple PBL outcomes with an agent.

%% file: sec/6.5_appendix.tex
\appendix
\section{Appendix} \label{appendix}
\textbf{The dataset contains potentially sensitive information, and IRB is currently reviewing PBL-STEM. The full dataset will be publicly released once the review process is completed and all ethical considerations are addressed.}
\subsection{More Experimental Analysis}\label{more_experimental}
\begin{figure}[h]
    \centering
    \includegraphics[width=\linewidth]{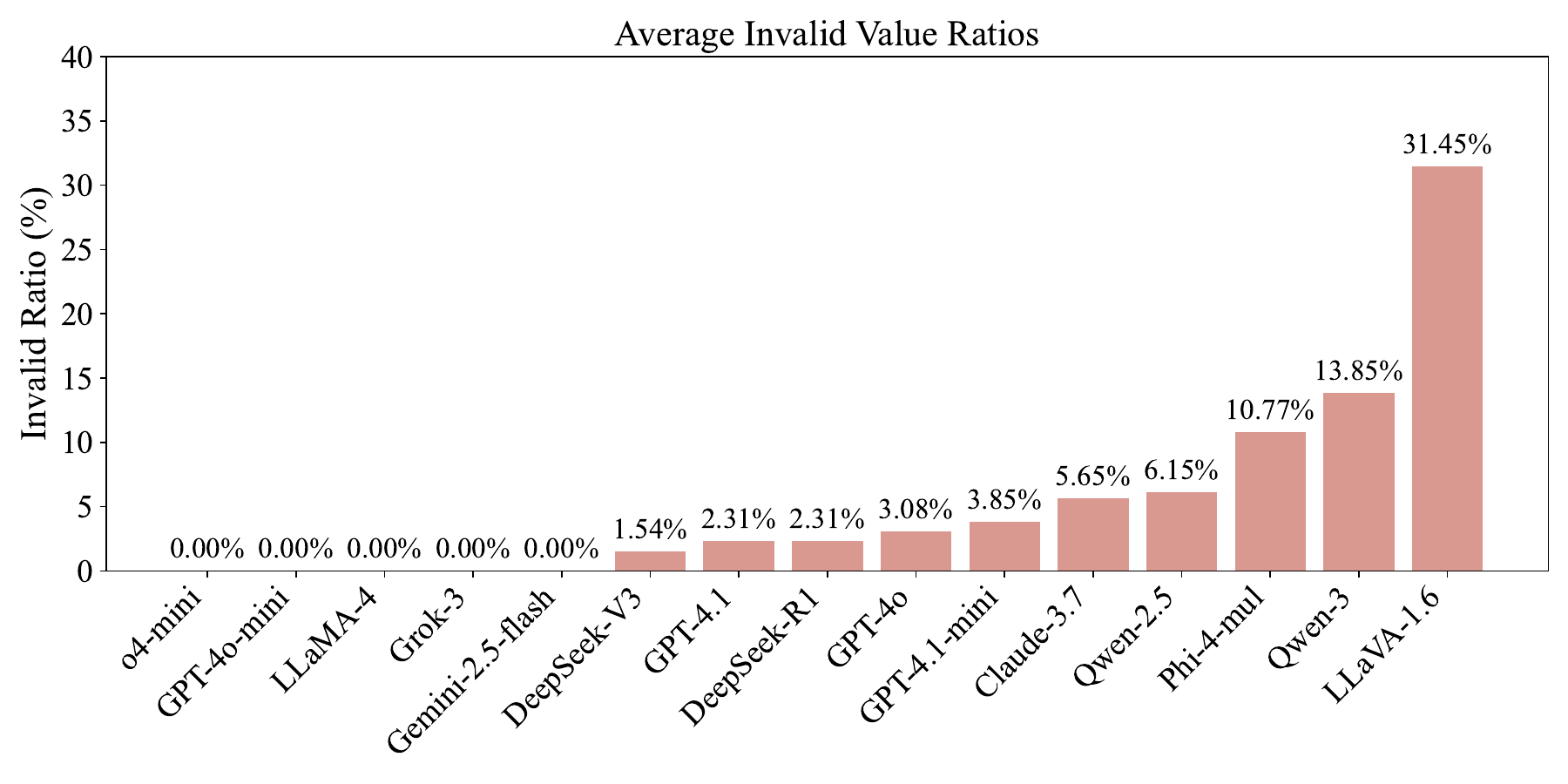}
    \caption{The comparison of \textbf{average invalid value ratios} for different models in PBLBench.}
    \label{fig:invalid}
\end{figure}

\textbf{Analysis of Invalid Value Ratios:}
Considering the capabilities of the model and issues with hallucinations, invalid values may be output during the evaluation of the project.
These invalid outputs are categorized as zero.
As shown in Figure \ref{fig:invalid}, we also present the proportion of invalid values output by the model during the evaluation.
It is not difficult to observe that models such as o4-mini and Gemini-2.5 are stable in performance, with a zero percent ratio of invalid outputs during all project evaluations.
In contrast, models like Qwen-3 and Phi-4 exhibit a higher proportion of invalid outputs, particularly the LLaVA-1.6 model, which has an invalid output ratio of 31.45\%. 

\begin{figure*}[h]
\vspace{-0.35\intextsep}
  \centering
  \captionsetup[subfloat]{font=scriptsize}
  \subfloat[Multiple projects]{\includegraphics[width=2.75in]{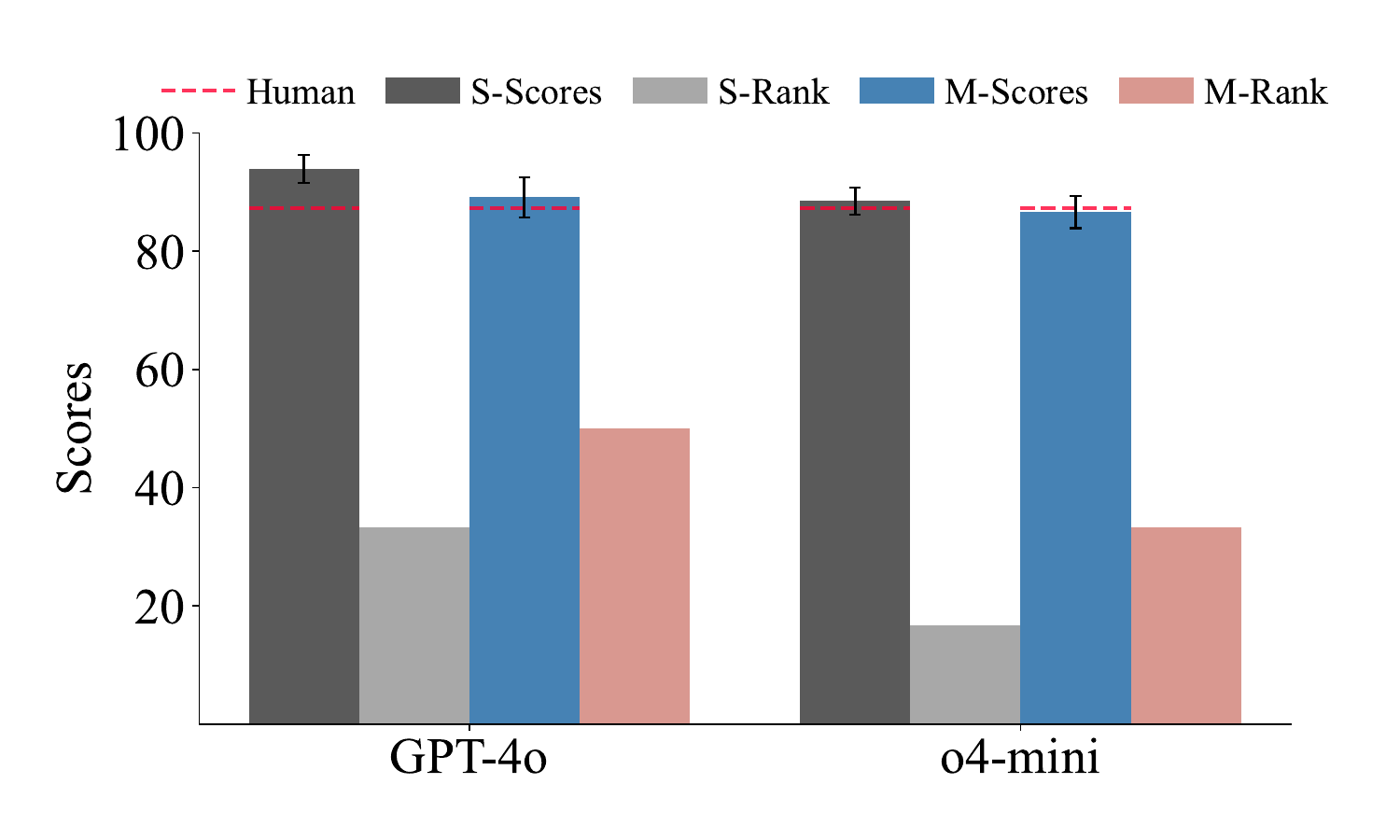}\label{fig:7.1}}
  \subfloat[Input order]{\includegraphics[width=2.75in]{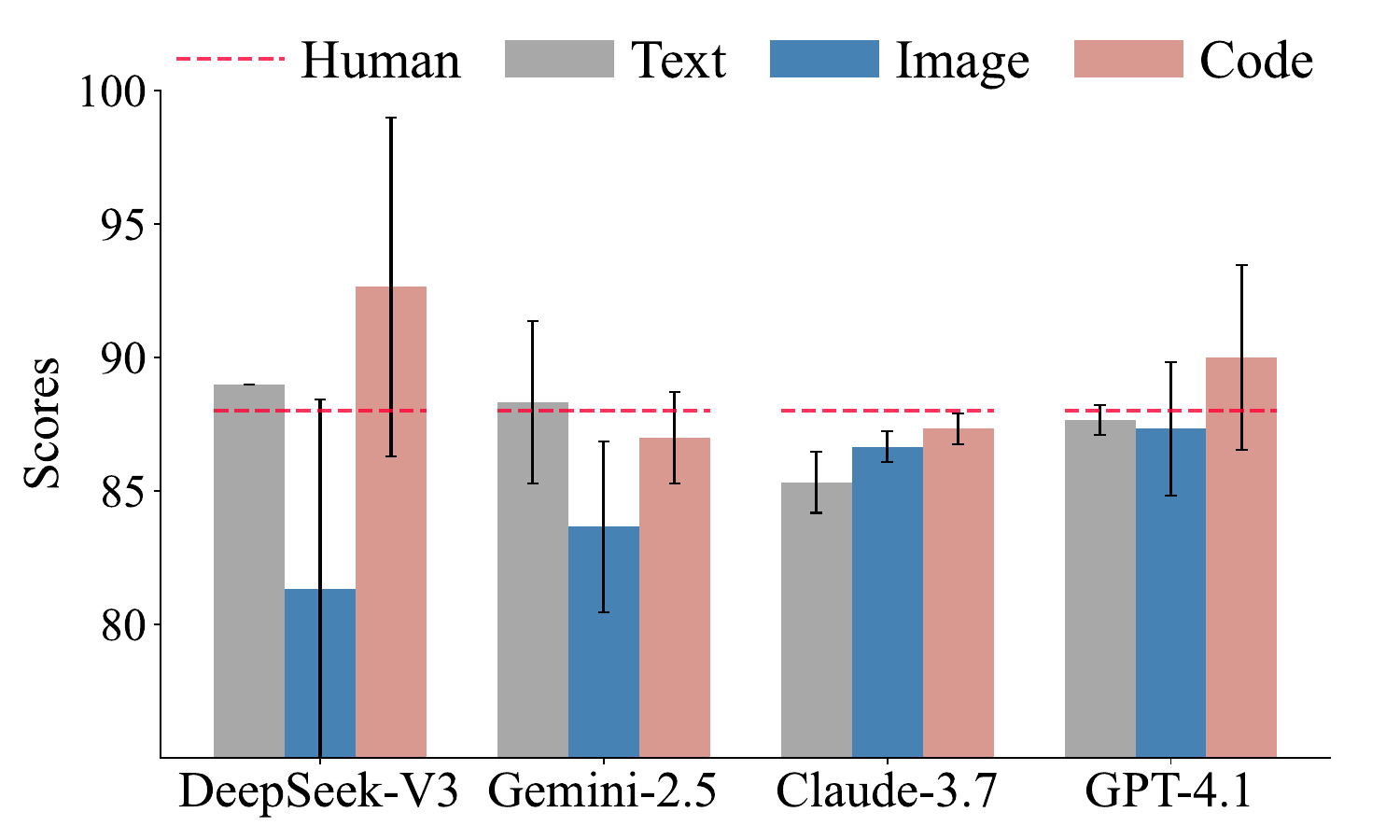}\label{fig:7.2}}
\vspace{-0.5\intextsep}
\caption{The performance comparison involves assessing a \textbf{single project versus multiple projects}, and \textbf{sequence of project material inputs}}
\vspace{-0.5\intextsep}
\label{figure:7} 
\end{figure*}

\textbf{Multi-project Evaluation Capability:}
Considering the limitation on input length, our experiments adopted the mode of evaluating a single project in one query. However, we also analyzed the capability of MLLMs to evaluate multiple projects within the input length limitations, as shown in Figure \ref{fig:7.1}. We found that the average scores of multi-document evaluations are similar to those of single-document evaluations, but the ranking accuracy is higher. For example, in the GPT-4o model, the ranking accuracy for multiple projects is close to 50\%. 

\textbf{Examining Variations in Input Orders:}
Previous research indicates that the order in which materials for different projects are input can also affect assessment performance~\cite{rafi2024impact}.
Therefore, we explored how different input orders of text, code, and images influence model scoring. The results are shown in Figure \ref{fig:7.2}.
It is not difficult to observe that in the DeepSeek-V3, Gemini-2.5, and GPT-4.1 models, using text as the initial input results in the most ideal scoring.
In settings where code is used as the initial input, the scores from the Claude-3.7 model more closely approximate human ratings and exhibit lower standard deviation.
Additionally, in the DeepSeek-V3 and GPT-4.1 models, the higher standard deviations in scores indicate greater instability.
Finally, in the GPT-4.1 and Claude-3.7 models, using image as the initial input results in better scoring compared to the DeepSeek-V3 and Gemini-2.5 models.

\begin{table*}[!t]
\centering
\setlength{\tabcolsep}{0.86mm}        
\vspace{-0.5\intextsep}
\caption{Performance comparison is based on different prompts in the discipline of science. "Prompt\_1" is used for our experiment, while "Prompt\_2" is leveraged for comparison.}
\vspace{-0.5\intextsep}
{{
\begin{tabular}{c|ccc|ccc|ccc|ccc}
    \toprule[1.5pt]
    \multirow{2}{*}{\textbf{Model}} & 
    \multicolumn{3}{c|}{\textbf{Qwen-2.5}} & 
    \multicolumn{3}{c|}{\textbf{DeepSeek-V3}} &  
    \multicolumn{3}{c|}{\textbf{GPT-4.1-mini}} &  
    \multicolumn{3}{c}{\textbf{GPT-4.1}} \\
    
\cmidrule(rl){2-4}\cmidrule(rl){5-7} \cmidrule(rl){8-10} \cmidrule(rl){11-13} 
& {Ave} & {Std} & {Acc} & {Ave} & {Std} & {Acc} & {Ave} & {Std} & {Acc} & {Ave} & {Std} & {Acc} \\
\hline
Prompt\_1   &88.53  &1.90   &15.78  &96.66  &3.42   &31.50   &89.99  &2.18   &42.10    &94.31 &1.15 &31.57 \\
Prompt\_2   &92.99	&2.07	&15.78  &92.99	&2.27	&26.31   &91.29	 &1.94	 &15.78    &75.01	 &1.36	& 21.05 \\
\toprule[1.5pt]
\end{tabular}}}
\vspace{-0.75\intextsep}
\label{tab_prompt}
\end{table*}

\textbf{Evaluation of Different Prompts:}
We also analyze the impact of different prompts on model evaluation; the prompt used in our experiments is presented in Tables \ref{tab_prompt1} and \ref{tab_prompt2}.
As shown in Table \ref{tab_prompt}, we observe that although prompt\_2 is closer to human scoring in the DeepSeek-V3 model, it has a lower ranking accuracy, which has decreased by 5.21\%.
Furthermore, in GPT-4.1, the scores for prompt\_1 are significantly better than those for prompt\_2, and are accompanied by a higher ranking accuracy.
The results above indicate that a detailed prompt enhances project evaluation.

In addition, we demonstrated the details of the dataset in Table \ref{tab_datset1}, which involves the proportion of samples from different disciplines, their lengths, and the modalities they possess.

\begin{figure}[!h]
    \centering
    \includegraphics[width=\linewidth]{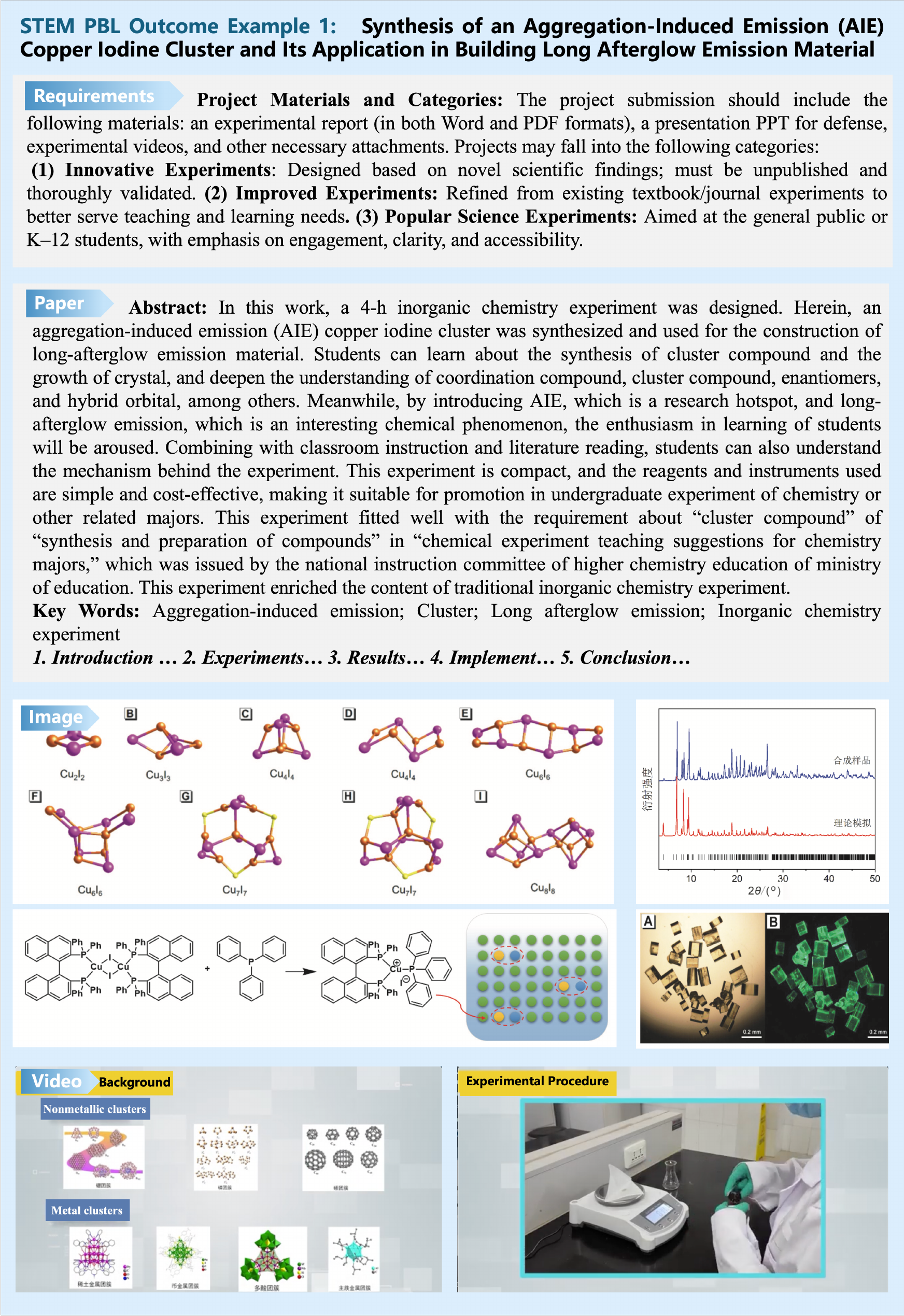}
    \caption{The visualized examples in science disciplines include reports, images, and video.}
    \label{fig:example1}
\end{figure}

\begin{figure}[!h]
    \centering
    \includegraphics[width=\linewidth]{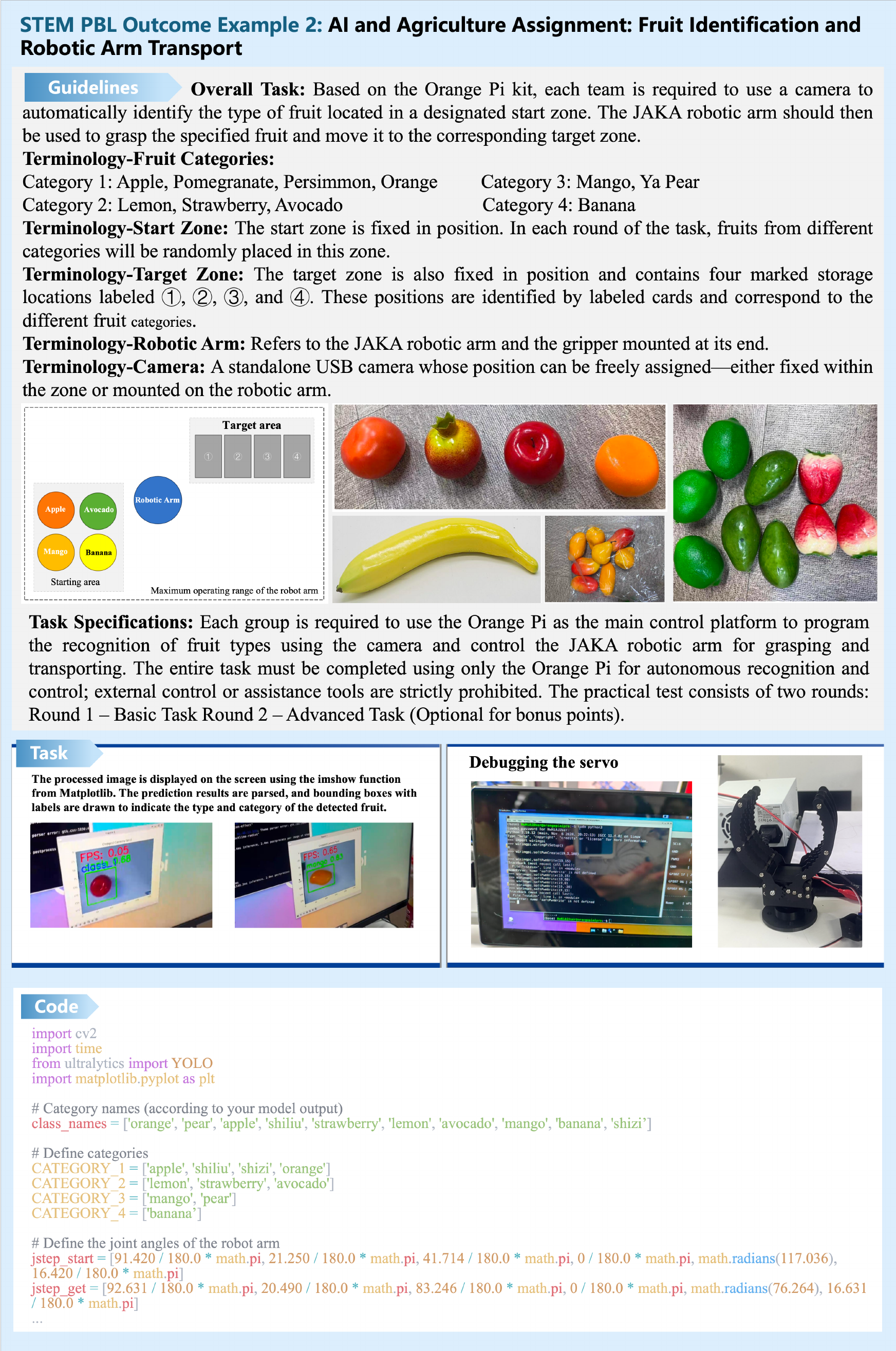}
    \caption{The visualized examples in technology disciplines include reports, images, and code.}
    \label{fig:example2}
\end{figure}

\begin{figure}[!h]
    \centering
    \includegraphics[width=\linewidth]{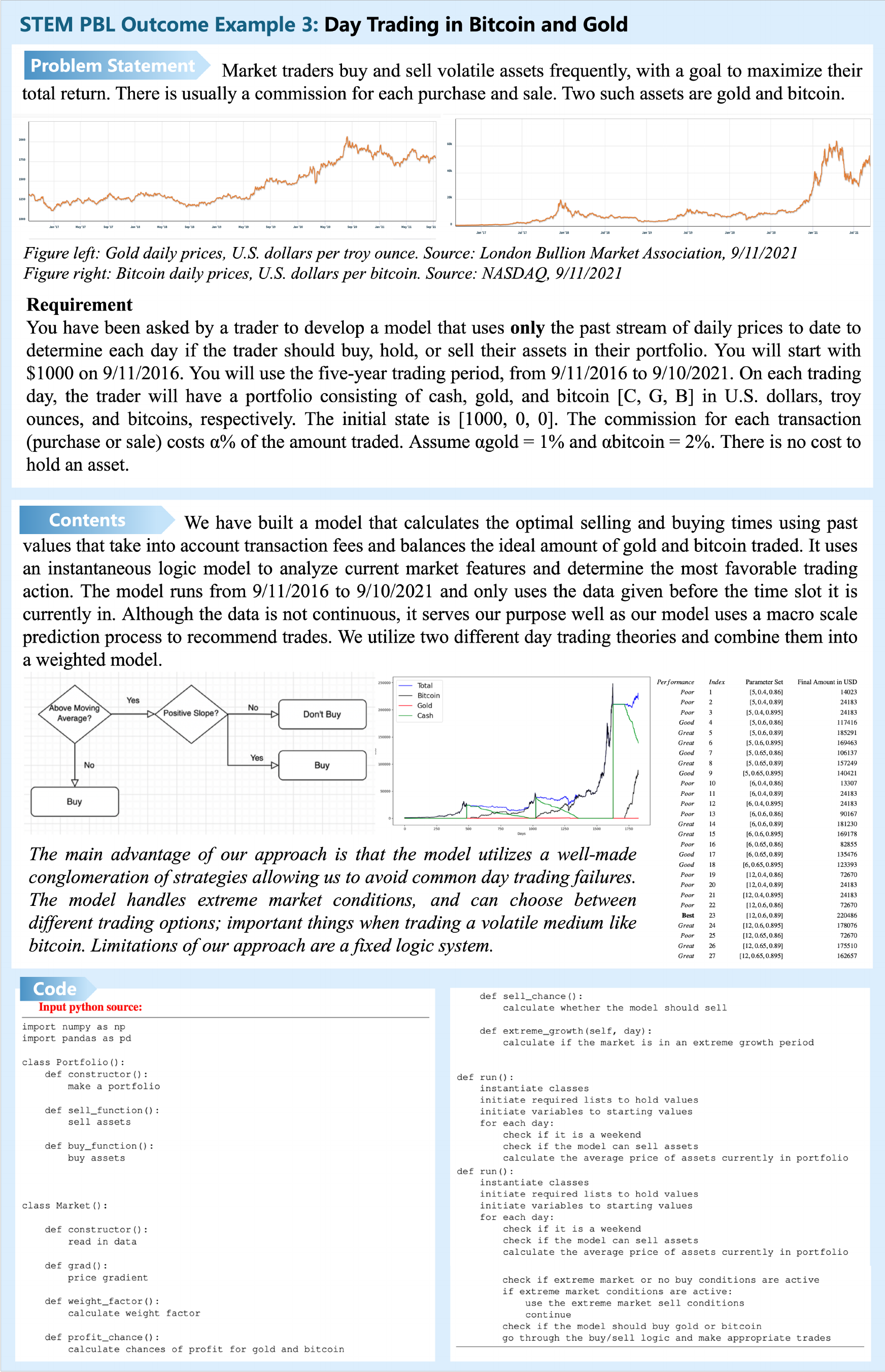}
    \caption{The visualized examples in mathematics disciplines include reports, images, and code.}
    \label{fig:example3}
\end{figure}

Finally, the model's reasoning process is demonstrated in Table \ref{tab_case_table}, and some visualizations of the project outcomes are shown in Figures \ref{fig:example1}, \ref{fig:example2}, and \ref{fig:example3}.

\begin{table}[ht]
	\centering
   \vspace{-3mm}
	\caption{The details of the PBL-STEM dataset include the sample distribution across each discipline, average project length, and modalities.}
	\setlength{\tabcolsep}{4.5pt} 
    \begin{tabular}{c|cccc}
        \bottomrule[1.2pt]
        Types         & Science & Technology &Engineering & Mathematics \\
        \bottomrule[1.2pt]
        Rate & 14.0\%  & 17.5\% & 15.4\%  & 53.1\% \\
        Length &  7228.45 & 3194.42 & 8473.24  & 11767.66 \\
        Text &  \cmark & \cmark & \cmark  & \cmark \\
        Image &  \cmark & \cmark & \cmark  & \cmark \\
        Code &  \xmark & \cmark & \cmark  & \cmark \\
        Video &  \cmark & \xmark & \cmark  & \xmark \\
        \bottomrule[1.2pt]
	\end{tabular}
    \label{tab_datset1}
\end{table}

\subsection{Limitations}
Although our PBLBench thoroughly evaluated the capabilities of state-of-the-art MLLMs in assessing PBL outcomes, we identify two major limitations of our work: 
(i) Despite the interdisciplinary scope of PBL-STEM, its overall scale remains limited, necessitating the collection of additional samples to enhance the comprehensiveness of the dataset.
(ii) Due to limitations in input length, we only evaluate one project per query, which results in a lack of comparison between projects.

\begin{table*}[htb]
\centering
\caption{The simplified prompt, which was leveraged as a comparison for the ablation study.}
\begin{tcolorbox}[width=0.99\linewidth, boxsep=0pt, left=2pt, right=2pt, top=1pt, bottom=0pt, colback=gray!10, after=\vspace{-3pt}]
\setlength{\baselineskip}{10pt}
\small You are a STEM education expert with a professional background tasked with evaluating students' multimodal project-based learning outcomes. Based on submitted documents, code, images, audio, etc., assess their performance in each dimension (with a focus on documentation). Evaluate the project in terms of knowledge, skill, expression, and innovation reflection, each with a score out of 25 (excellent, good, average, poor). Consider possible biases or gaps in the evaluation. Finally, output a project assessment of no more than 100 words with a total score out of 100 (only output an integer for the final score). \strut 
\end{tcolorbox}
\label{tab_prompt2}
\end{table*}

\begin{table*}[htb]
\centering
\caption{The detailed prompt used in PBLBench, which was written by humans following the evaluation criteria.}
\begin{tcolorbox}[width=0.99\linewidth, boxsep=0pt, left=2pt, right=2pt, top=1pt, bottom=0pt, colback=gray!10, after=\vspace{-3pt}]
\setlength{\baselineskip}{10pt}
\small You are a STEM education expert with a professional background, required to assess the project-based learning outcomes submitted by students. The overall score for the project-based learning outcomes is calculated using a weighted composite of four core dimensions, with a perfect score of 100. \strut 

{\bf \small 1. Knowledge dimension includes:} 
\small  1) Concept understanding (15 points); 2) Interdisciplinary application (5 points); 3) Evidence-based reasoning (5 points). \strut 

{\bf \small 2. Skills dimension includes:}
\small 1) Tool and process application (10 points); 2) Problem-solving ability (10 points); 3) Self-regulation and planning (10 points).  \strut 

{\bf \small 3. Expression dimension includes:}
\small 1) Clarity of information expression (8 points); 2) Multimodal expression capability (6 points); 3) Audience awareness and adaptation (6 points). \strut 

{\bf \small 4. Innovation and reflection dimension includes:}
\small 1) Innovation and practicality (15 points); 2) Creative development and iteration (5 points); 3) Self-reflection and growth (5 points).  \strut 

\small Your task: For each of the 12 sub-indicators listed above, provide: the corresponding score for each sub-indicator and a brief reason for the score (based on content quality, logic, completeness, etc.). Finally, give the total score for the entire project (output a single integer). \strut 

\end{tcolorbox}
\label{tab_prompt1}
\end{table*}

\begin{table*}[htb]
\centering
\caption{The case study for the reasoning process of the Gemini-2.5 model}
\begin{tcolorbox}[width=0.99\linewidth, boxsep=0pt, left=2pt, right=2pt, top=1pt, bottom=0pt, colback=gray!10, after=\vspace{-3pt}]
\setlength{\baselineskip}{10pt}
{\bf \small After multiple evaluations, the performance of the project in the following aspects is as follows:} \strut 

{\bf \small Conceptual Understanding (15 points):} \strut 

\small Score: 14 points; Reason: The project demonstrates a solid understanding of the basic concepts of fruit recognition and robotic arm operation, and it provides a detailed explanation of the fruit classification and processing workflow. \strut 

{\bf \small Interdisciplinary Application (5 points):} \strut 

\small Score: 4 points; Reason: The student demonstrated an integrated application of disciplines such as computer vision, machine learning, and mechanical control, but the explanation of applications in physics or engineering was somewhat lacking. \strut 

{\bf \small Evidence-Based Reasoning (5 points):} \strut 

\small Score: 4 points; Reason: The basic evidence-based reasoning capability is demonstrated through model training and prediction results, but there is a lack of in-depth discussion on model performance metrics. \strut 

{\bf \small Tools and Process Application (10 points):} \strut 

\small Score: 9 points; Reason: The project demonstrates proficient use of toolkits like OpenCV and TensorFlow, with a logical and complete process design; however, there is room for simplification of redundant code. \strut 

{\bf \small Problem-Solving Ability (10 points):} \strut 

\small Score: 8 points; Reason: Effectively resolved key issues in fruit recognition and robotic arm control, but the project's robustness against various disturbances needs improvement, as shown by the code and documentation.\strut 

{\bf \small Autonomous Regulation and Planning (10 points):} \strut

\small Score: 9 points; Reason: The project plan was executed in an orderly manner, capable of autonomously controlling the execution of work tasks, but further exploration is needed in handling more challenging tasks. \strut 

{\bf \small Clarity of Information Expression (8 points):} \strut 

\small Score: 7 points; Reason: The document structure is clear and the information is fairly complete, but the explanation of some details is somewhat vague. \strut 

{\bf \small Multimodal Expression Ability (6 points):} \strut 

\small Score: 5 points; Reason: The project utilizes multimodal expression methods such as visual images and code annotations, but lacks in areas like sound and real-time interactive demonstrations. \strut 

{\bf \small Audience Awareness and Adaptation (6 points):} \strut 

\small Score: 5 points; Reason: The target audience includes judges and classmates, and the materials prepared are fairly appropriate, but the presentation materials such as PPT are somewhat simplistic. \strut 

{\bf \small Innovativeness and Practicality (15 points):} \strut 

\small Score: 13 points; Reason: The project demonstrates innovative applications based on Orange Pi, which have certain practicality, but the technological innovations could be further explored. \strut 

{\bf \small Creativity Development and Iteration (5 points):} \strut 

\small Score: 4 points; Reason: Reflects basic project creativity and evolution, but lacks sufficient documentation of creative updates and project iterative improvements. \strut 

{\bf \small Self-reflection and Growth (5 points):} \strut 

\small Score: 3 points; Reason: The project documentation lacks reflection on personal growth and challenges encountered during the project process. \strut 

{\bf \small Based on the above assessment, the final score is 85 points.} \strut 
\end{tcolorbox}

\label{tab_case_table}
\end{table*}

\clearpage


%% file: main.bbl
\begin{thebibliography}{10}

\bibitem{abdin2024phi}
Marah Abdin, Jyoti Aneja, Harkirat Behl, S{\'e}bastien Bubeck, Ronen Eldan, Suriya Gunasekar, Michael Harrison, Russell~J Hewett, Mojan Javaheripi, Piero Kauffmann, et~al.
\newblock Phi-4 technical report.
\newblock {\em arXiv preprint arXiv:2412.08905}, 2024.

\bibitem{Claude3S}
Anthropic.
\newblock Claude 3.7 sonnet system card.

\bibitem{austin2021program}
Jacob Austin, Augustus Odena, Maxwell Nye, Maarten Bosma, Henryk Michalewski, David Dohan, Ellen Jiang, Carrie Cai, Michael Terry, Quoc Le, et~al.
\newblock Program synthesis with large language models.
\newblock {\em arXiv preprint arXiv:2108.07732}, 2021.

\bibitem{chen2024restoreagent}
Haoyu Chen, Wenbo Li, Jinjin Gu, Jingjing Ren, Sixiang Chen, Tian Ye, Renjing Pei, Kaiwen Zhou, Fenglong Song, and Lei Zhu.
\newblock Restoreagent: Autonomous image restoration agent via multimodal large language models.
\newblock {\em arXiv preprint arXiv:2407.18035}, 2024.

\bibitem{chen2021codex}
Mark Chen, Jerry Tworek, Heewoo Jun, Qiming Yuan, Henrique~Ponde de~Oliveira~Pinto, Jared Kaplan, Harri Edwards, Yuri Burda, Nicholas Joseph, Greg Brockman, Alex Ray, Raul Puri, Gretchen Krueger, Michael Petrov, Heidy Khlaaf, Girish Sastry, Pamela Mishkin, Brooke Chan, Scott Gray, Nick Ryder, Mikhail Pavlov, Alethea Power, Lukasz Kaiser, Mohammad Bavarian, Clemens Winter, Philippe Tillet, Felipe~Petroski Such, Dave Cummings, Matthias Plappert, Fotios Chantzis, Elizabeth Barnes, Ariel Herbert-Voss, William~Hebgen Guss, Alex Nichol, Alex Paino, Nikolas Tezak, Jie Tang, Igor Babuschkin, Suchir Balaji, Shantanu Jain, William Saunders, Christopher Hesse, Andrew~N. Carr, Jan Leike, Josh Achiam, Vedant Misra, Evan Morikawa, Alec Radford, Matthew Knight, Miles Brundage, Mira Murati, Katie Mayer, Peter Welinder, Bob McGrew, Dario Amodei, Sam McCandlish, Ilya Sutskever, and Wojciech Zaremba.
\newblock Evaluating large language models trained on code.
\newblock 2021.

\bibitem{chou2003redefining}
Chih-Yueh Chou, Tak-Wai Chan, and Chi-Jen Lin.
\newblock Redefining the learning companion: the past, present, and future of educational agents.
\newblock {\em Computers \& Education}, 40(3):255--269, 2003.

\bibitem{cobbe2021training}
Karl Cobbe, Vineet Kosaraju, Mohammad Bavarian, Mark Chen, Heewoo Jun, Lukasz Kaiser, Matthias Plappert, Jerry Tworek, Jacob Hilton, Reiichiro Nakano, et~al.
\newblock Training verifiers to solve math word problems.
\newblock {\em arXiv preprint arXiv:2110.14168}, 2021.

\bibitem{ewis2023assessment}
Ammar Ewis and Rasha El~Shafie.
\newblock Assessment of a project-based learning versus conventional engineering program using analytical hierarchy technique.
\newblock {\em Journal of Advanced Education and Sciences}, 3(1):65--74, 2023.

\bibitem{firdausih2024literature}
Firdausih Firdausih and Aslan Aslan.
\newblock Literature review: The effect of project-based learning on student motivation and achievement in science.
\newblock {\em Indonesian Journal of Education (INJOE)}, 4(3):1011--1022, 2024.

\bibitem{gao12024check}
Chenglin Gao, Yixuan Jiang, and Tao Yu.
\newblock The application of analytic hierarchy process (ahp) in project-based teaching practice.
\newblock In {\em Proceedings of the 4th International Conference on New Computational Social Science (ICNCSS 2024)}, page~4. Springer Nature, 2024.

\bibitem{google-gemini2.5-flash-preview-2025}
{Google}.
\newblock {Gemini 2.5 Flash Preview Model Card}.
\newblock \url{https://storage.googleapis.com/model-cards/documents/gemini-2.5-flash-preview.pdf}, 2025.

\bibitem{guo2025deepseek}
Daya Guo, Dejian Yang, Haowei Zhang, Junxiao Song, Ruoyu Zhang, Runxin Xu, Qihao Zhu, Shirong Ma, Peiyi Wang, Xiao Bi, et~al.
\newblock Deepseek-r1: Incentivizing reasoning capability in llms via reinforcement learning.
\newblock {\em arXiv preprint arXiv:2501.12948}, 2025.

\bibitem{hendrycks2024measuring}
Dan Hendrycks, Collin Burns, Steven Basart, Andy Zou, Mantas Mazeika, Dawn Song, and Jacob Steinhardt.
\newblock Measuring massive multitask language understanding.
\newblock In {\em International Conference on Learning Representations}, 2021.

\bibitem{hendrycks2021measuring}
Dan Hendrycks, Collin Burns, Saurav Kadavath, Akul Arora, Steven Basart, Eric Tang, Dawn Song, and Jacob Steinhardt.
\newblock Measuring mathematical problem solving with the math dataset.
\newblock In {\em Thirty-fifth Conference on Neural Information Processing Systems Datasets and Benchmarks Track (Round 2)}, 2021.

\bibitem{hwang2020vision}
Gwo-Jen Hwang, Haoran Xie, Benjamin~W Wah, and Dragan Ga{\v{s}}evi{\'c}.
\newblock Vision, challenges, roles and research issues of artificial intelligence in education, 2020.

\bibitem{jia2025uni}
Yanhao Jia, Xinyi Wu, Hao Li, Qinglin Zhang, Yuxiao Hu, Shuai Zhao, and Wenqi Fan.
\newblock Uni-retrieval: A multi-style retrieval framework for stem's education.
\newblock {\em arXiv preprint arXiv:2502.05863}, 2025.

\bibitem{kim2024mdagents}
Yubin Kim, Chanwoo Park, Hyewon Jeong, Yik~S Chan, Xuhai Xu, Daniel McDuff, Hyeonhoon Lee, Marzyeh Ghassemi, Cynthia Breazeal, and Hae~W Park.
\newblock Mdagents: An adaptive collaboration of llms for medical decision-making.
\newblock {\em Advances in Neural Information Processing Systems}, 37:79410--79452, 2024.

\bibitem{free}
Hao Li, Yanhao Jia, Jin Peng, Zesen Cheng, Kehan Li, Jialu Sui, Chang Liu, and Li~Yuan.
\newblock Freestyleret: Retrieving images from style-diversified queries.
\newblock In {\em Computer Vision -- ECCV 2024}, pages 258--274, Cham, 2025. Springer Nature Switzerland.

\bibitem{li2024alleviating}
Tieying Li, Xiaochun Yang, Yiping Ke, Bin Wang, Yinan Liu, and Jiaxing Xu.
\newblock Alleviating the inconsistency of multimodal data in cross-modal retrieval.
\newblock In {\em 2024 IEEE 40th International Conference on Data Engineering (ICDE)}, pages 4643--4656. IEEE, 2024.

\bibitem{liu2024deepseek}
Aixin Liu, Bei Feng, Bing Xue, Bingxuan Wang, Bochao Wu, Chengda Lu, Chenggang Zhao, Chengqi Deng, Chenyu Zhang, Chong Ruan, et~al.
\newblock Deepseek-v3 technical report.
\newblock {\em arXiv preprint arXiv:2412.19437}, 2024.

\bibitem{liu2023improved}
Haotian Liu, Chunyuan Li, Yuheng Li, and Yong~Jae Lee.
\newblock Improved baselines with visual instruction tuning, 2023.

\bibitem{lu2022learn}
Pan Lu, Swaroop Mishra, Tanglin Xia, Liang Qiu, Kai-Wei Chang, Song-Chun Zhu, Oyvind Tafjord, Peter Clark, and Ashwin Kalyan.
\newblock Learn to explain: Multimodal reasoning via thought chains for science question answering.
\newblock {\em Advances in Neural Information Processing Systems}, 35:2507--2521, 2022.

\bibitem{meta2025llama}
AI~Meta.
\newblock The llama 4 herd: The beginning of a new era of natively multimodal ai innovation.
\newblock {\em https://ai. meta. com/blog/llama-4-multimodal-intelligence/, checked on}, 2025.

\bibitem{nguyen2024kdmcse}
Cong-Duy Nguyen, Thong Nguyen, Xiaobao Wu, and Luu~Anh Tuan.
\newblock Kdmcse: Knowledge distillation multimodal sentence embeddings with adaptive angular margin contrastive learning.
\newblock In {\em Proceedings of the 2024 Conference of the North American Chapter of the Association for Computational Linguistics: Human Language Technologies (Volume 1: Long Papers)}, pages 733--749, 2024.

\bibitem{nguyen2025enhancing}
Cong-Duy Nguyen, Xiaobao Wu, Thong Nguyen, Shuai Zhao, Khoi Le, Viet-Anh Nguyen, Feng Yichao, and Anh~Tuan Luu.
\newblock Enhancing multimodal entity linking with jaccard distance-based conditional contrastive learning and contextual visual augmentation.
\newblock {\em arXiv preprint arXiv:2501.14166}, 2025.

\bibitem{openai_gpt4_1}
OpenAI.
\newblock Gpt-4.1 overview.
\newblock \url{https://openai.com/index/gpt-4-1/}, 2025.

\bibitem{openai_gpt4o_mini}
OpenAI.
\newblock Gpt-4o mini: Advancing cost-efficient intelligence.
\newblock \url{https://openai.com/index/gpt-4o-mini-advancing-cost-efficient-intelligence/}, 2025.

\bibitem{openai-o3-o4-mini-system-card-2025}
{OpenAI}.
\newblock {OpenAI o3 and o4-mini System Card}.
\newblock \url{https://openai.com/index/o3-o4-mini-system-card/}, March 2025.

\bibitem{pichai2024introducing}
Sundar Pichai, D~Hassabis, and K~Kavukcuoglu.
\newblock Introducing gemini 2.0: our new ai model for the agentic era, 2024.

\bibitem{qwen3-2025}
{QwenLM Team}.
\newblock {Qwen3: Think Deeper, Act Faster}.
\newblock \url{https://qwenlm.github.io/blog/qwen3/}, 2025.

\bibitem{rafi2024impact}
Md~Nakhla Rafi, Dong~Jae Kim, Tse-Hsun Chen, and Shaowei Wang.
\newblock The impact of input order bias on large language models for software fault localization.
\newblock {\em arXiv preprint arXiv:2412.18750}, 2024.

\bibitem{shen2024measuring}
Jianhao Shen, Ye~Yuan, Srbuhi Mirzoyan, Ming Zhang, and Chenguang Wang.
\newblock Measuring vision-language stem skills of neural models.
\newblock In {\em ICLR}, 2024.

\bibitem{viswanathan2022enhancement}
Nethra Viswanathan, Sofia Meacham, and Festus~Fatai Adedoyin.
\newblock Enhancement of online education system by using a multi-agent approach.
\newblock {\em Computers and Education: Artificial Intelligence}, 3:100057, 2022.

\bibitem{wang2024novelqa}
Cunxiang Wang, Ruoxi Ning, Boqi Pan, Tonghui Wu, Qipeng Guo, Cheng Deng, Guangsheng Bao, Xiangkun Hu, Zheng Zhang, Qian Wang, et~al.
\newblock Novelqa: Benchmarking question answering on documents exceeding 200k tokens.
\newblock {\em arXiv preprint arXiv:2403.12766}, 2024.

\bibitem{xai-grok3-2024}
{xAI}.
\newblock {Grok 3 Beta — The Age of Reasoning Agents}.
\newblock \url{https://x.ai/news/grok-3}, 2025.

\bibitem{xiao}
Luwei Xiao, Rui Mao, Shuai Zhao, Qika Lin, Yanhao Jia, Liang He, and Erik Cambria.
\newblock Exploring cognitive and aesthetic causality for multimodal aspect-based sentiment analysis.
\newblock {\em IEEE Transactions on Affective Computing}, pages 1--18, 2025.

\bibitem{xu2024eduagent}
Songlin Xu, Xinyu Zhang, and Lianhui Qin.
\newblock Eduagent: Generative student agents in learning.
\newblock {\em arXiv preprint arXiv:2404.07963}, 2024.

\bibitem{yang2024qwen2}
An~Yang, Baosong Yang, Beichen Zhang, Binyuan Hui, Bo~Zheng, Bowen Yu, Chengyuan Li, Dayiheng Liu, Fei Huang, Haoran Wei, et~al.
\newblock Qwen2. 5 technical report.
\newblock {\em arXiv preprint arXiv:2412.15115}, 2024.

\bibitem{yu2024fincon}
Yangyang Yu, Zhiyuan Yao, Haohang Li, Zhiyang Deng, Yuechen Jiang, Yupeng Cao, Zhi Chen, Jordan Suchow, Zhenyu Cui, Rong Liu, et~al.
\newblock Fincon: A synthesized llm multi-agent system with conceptual verbal reinforcement for enhanced financial decision making.
\newblock {\em Advances in Neural Information Processing Systems}, 37:137010--137045, 2024.

\bibitem{yue2023mmmu}
Xiang Yue, Yuansheng Ni, Kai Zhang, Tianyu Zheng, Ruoqi Liu, Ge~Zhang, Samuel Stevens, Dongfu Jiang, Weiming Ren, Yuxuan Sun, Cong Wei, Botao Yu, Ruibin Yuan, Renliang Sun, Ming Yin, Boyuan Zheng, Zhenzhu Yang, Yibo Liu, Wenhao Huang, Huan Sun, Yu~Su, and Wenhu Chen.
\newblock Mmmu: A massive multi-discipline multimodal understanding and reasoning benchmark for expert agi.
\newblock In {\em Proceedings of CVPR}, 2024.

\bibitem{zhao2024weak}
Shuai Zhao, Leilei Gan, Zhongliang Guo, Xiaobao Wu, Luwei Xiao, Xiaoyu Xu, Cong-Duy Nguyen, and Luu~Anh Tuan.
\newblock Weak-to-strong backdoor attack for large language models.
\newblock {\em arXiv preprint arXiv:2409.17946}, 2024.

\bibitem{zhao2024uni}
Shuai Zhao, Meihuizi Jia, Luu~Anh Tuan, Fengjun Pan, and Jinming Wen.
\newblock Universal vulnerabilities in large language models: Backdoor attacks for in-context learning.
\newblock In {\em Proceedings of the 2024 Conference on Empirical Methods in Natural Language Processing}, pages 11507--11522, 2024.

\bibitem{zhao2024unlearning}
Shuai Zhao, Xiaobao Wu, Cong-Duy Nguyen, Meihuizi Jia, Yichao Feng, and Luu~Anh Tuan.
\newblock Unlearning backdoor attacks for llms with weak-to-strong knowledge distillation.
\newblock {\em arXiv preprint arXiv:2410.14425}, 2024.

\bibitem{zoudynamath}
Chengke Zou, Xingang Guo, Rui Yang, Junyu Zhang, Bin Hu, and Huan Zhang.
\newblock Dynamath: A dynamic visual benchmark for evaluating mathematical reasoning robustness of vision language models.
\newblock In {\em The Thirteenth International Conference on Learning Representations}, 2025.

\end{thebibliography}
